\documentclass{article}

\usepackage{iclr2026_conference,times}

\usepackage{amsmath,amsfonts,bm}

\def\eqref#1{equation~\ref{#1}}

\def\1{\bm{1}}

\DeclareMathAlphabet{\mathsfit}{\encodingdefault}{\sfdefault}{m}{sl}
\SetMathAlphabet{\mathsfit}{bold}{\encodingdefault}{\sfdefault}{bx}{n}

\iclrfinalcopy

\usepackage{xspace}
\usepackage{amsmath,amsfonts}
\usepackage{graphicx}
\usepackage{booktabs}
\usepackage{nicefrac}
\usepackage{microtype}
\usepackage{ragged2e}
\usepackage{placeins}
\usepackage{wrapfig}
\usepackage{tabularx}
\usepackage[table]{xcolor}
\usepackage{siunitx}
\usepackage{makecell}
\usepackage{subcaption}
\usepackage{enumitem}
\usepackage{bbm}
\usepackage{url}
\usepackage{pifont}
\usepackage{multirow}

\usepackage{listings}
\AtBeginDocument{}

\definecolor{citecolor}{RGB}{66,168,235}
\definecolor{linkcolor}{RGB}{255,0,0}
\usepackage{hyperref}
\hypersetup{colorlinks=true,citecolor=citecolor,linkcolor=linkcolor,urlcolor=citecolor}

\newcommand{\eg}{\emph{e.g.}\xspace} \newcommand{\ie}{\emph{i.e.}\xspace}   \newcommand{\cf}{\emph{cf.}\xspace} \newcommand{\mvs}{\emph{vs.}\xspace}

\newcommand{\drop}[1]{\textcolor{gray}{\scriptsize~(#1\%\,$\downarrow$)}}

\newcommand{\model}{\texttt{DyME}\xspace}

\newcommand{\raa}[1]{\renewcommand{\arraystretch}{#1}}

\title{Empowering Small VLMs to Think with Dynamic Memorization and Exploration}

\definecolor{darkgreen}{RGB}{0,100,0}
\usepackage{xcolor}

\usepackage{orcidlink}

\author{Jiazhen Liu\orcidlink{0000-0003-0584-4571}, Yuchuan Deng\orcidlink{0009-0009-0537-6801}, and Long Chen\thanks{Corresponding author (longchen@ust.hk)}\\
The Hong Kong University of Science and Technology \\ 
\tt\small\href{https://github.com/HKUST-LongGroup/DyME}{https://github.com/HKUST-LongGroup/DyME}
}

\begin{document}

\maketitle

\begin{abstract}
Small-scale Vision--Language Models (SVLMs) are exceptionally well-suited for proprietary tasks. Equipping them with thinking capabilities is a critical step to enhance their performance and reliability in these specific domains. However, existing training paradigms, including Supervised Fine-Tuning (SFT) and Reinforcement Learning with Verifiable Reward (RLVR), impose substantial demands on the base VLM, exceeding the capacity of SVLMs. Consequently, directly applying these paradigms to SVLMs fails to instill the desired thinking abilities.
A natural solution is to combine SFT and RLVR, leveraging their complementarity to reduce the dependence on model capacity. Yet the core challenge lies in managing the inherent trade-off: excessive reliance on SFT can force the model to memorize pseudo thinking traces, while over-emphasizing RLVR can lead to unstable exploration (\ie, advantage collapse). 
To address this, we propose \model, a novel training paradigm that \textbf{Dy}namically selects between \textbf{M}emorization (via SFT) and \textbf{E}xploration (via RLVR) at each optimization step. {By ensuring that every update contributes to the trade-off, \model serves as a robust, standalone strategy that stabilizes SVLM learning. Complementing this paradigm, we further introduce a synergistic \textit{Visual Supervision} mechanism (comprising a visual checker and refiner) designed to inject dynamically enhanced, image-grounded guidance during optimization.}
Extensive experiments across diverse domains demonstrate that \model consistently achieves this balance, and thus delivers substantial performance improvements on specialized tasks. These results establish \model as a practical and effective solution for empowering SVLMs with reliable thinking capabilities.
\end{abstract}


\section{Introduction}
Equipping Vision--Language Models (VLMs) with thinking capabilities is a pivotal step that moves them beyond recognition toward reasoning. Recent studies have advanced this goal through specialized training, achieving strong results on a spectrum of visual tasks, from recognition-intensive applications like grounding~\citep{lai2025med, liu2025segmentation, lmmr1, vrft, liu2025better} to reasoning-intensive challenges such as chart understanding~\citep{r1vl, xia2024chartx} and geometric problem solving~\citep{vlmr1, chen2025r1v, xia2024geox}. While this progress is significant, the success of these approaches is contingent upon the base VLM possessing strong foundational capabilities, namely, sufficient capacity and robust instruction adherence~\citep{GRPO4llava}. In practice, only a handful of VLMs meet these prerequisites, presenting a significant challenge for Small-scale VLMs (SVLMs) which struggle to develop thinking capabilities under existing training paradigms.

\begin{figure}[t]
  \centering
  \includegraphics[width=1\linewidth]{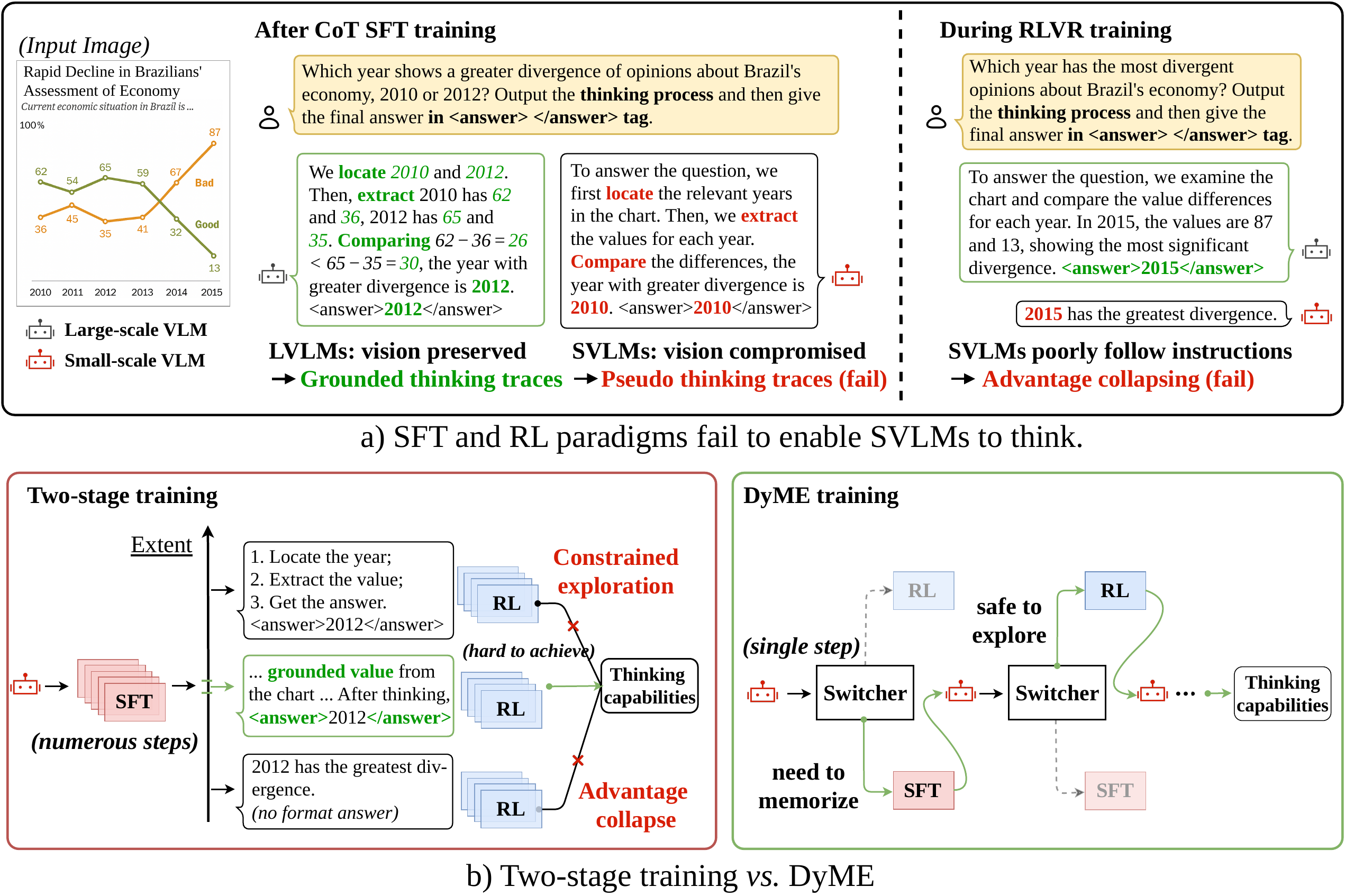}
\caption{\textbf{Training paradigms for enabling VLM thinking}. The LVLM is Qwen2.5-VL-32B~\citep{Qwen2.5-VL} and the SVLM is SmolVLM-500M~\citep{marafioti2025smolvlm}.
(a) Existing paradigms are effective for LVLMs but unsuitable for SVLMs. 
(b) The two-stage training paradigm (SFT $\rightarrow$ RL) faces a challenging trade-off. Our proposed \model dynamically balances this trade-off.}
  \label{fig:fig1}
\end{figure}

To contextualize this limitation, we briefly review the two dominant paradigms, both of which are primarily tailored for Large-scale VLMs (LVLMs).
\textbf{1) Supervised Fine-Tuning (SFT) on Chain-of-Thought (CoT) data}~\citep{xu2024llavacot, li2024synthesize, xia2024geox, gao2023gllava}: 
VLMs are supervised to memorize predefined thinking patterns from large-scale CoT annotations. Since CoT data are often verbose and contain much vision-irrelevant content, models must possess sufficient capacity to absorb long textual content without compromising visual grounding~\citep{marafioti2025smolvlm}. This capability gap is illustrated in Fig.~\ref{fig:fig1}a: After SFT, LVLMs can generate grounded thinking traces with accurate intermediate values (in green), while SVLMs cannot.
\textbf{2) Reinforcement Learning with Verifiable Reward (RLVR)}~\citep{r1vl, chen2025r1v, lmmr1, vlmr1}: on the other hand, promotes exploration of thinking patterns rather than imitations. 
In this paradigm, VLMs are instructed to generate a thought process followed by a strictly formatted answer (\eg, enclosed in tags). This format enables verifiable rewards to reinforce correct generations and penalize incorrect ones. Owing to its reliance on instruction adherence, this approach is practical primarily for strong VLMs that can reliably generate structured outputs.

\begin{wrapfigure}{r}{0.4\linewidth}  
    \centering
    \includegraphics[width=\linewidth]{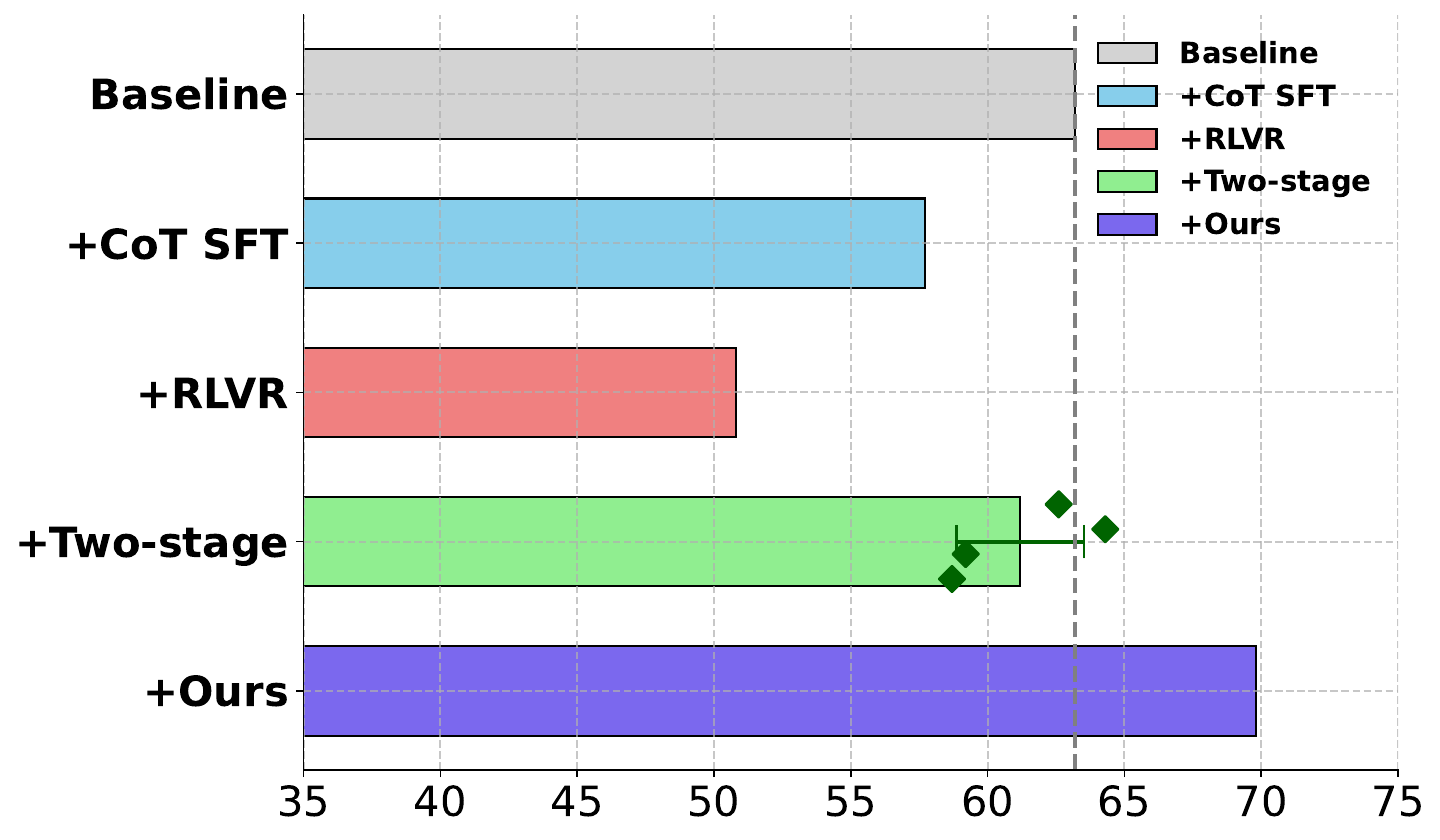}
    \caption{\textbf{Performance of SmolVLM-500M~\citep{marafioti2025smolvlm} on ChartQA ~\citep{masry-etal-2022-chartqa}.} Existing paradigms degrade performance, whereas \model yields improvements.}

    \label{fig:intro_demo}
\end{wrapfigure}

Consequently, both established paradigms are inadequate for instilling thinking in SVLMs. The extremely limited capacity (\eg, under 1B parameters) of SVLMs renders the SFT paradigm ineffective, as a high volume of textual information in CoT data can overwhelm the capacity~\citep{marafioti2025smolvlm, chen2025sft}.
Moreover, the limited instruction adherence of SVLMs frequently results in unverifiable outputs~\citep{chu2025sftmrl, guo2025deepseek}, precipitating advantage collapse during RLVR. We quantitatively verify these limitations (\cf, Fig.~\ref{fig:intro_demo}): both SFT and RLVR paradigms indeed impair the performance.

Considering that SVLMs offer high efficiency and are crucial for deployment on edge devices~\citep{marafioti2025smolvlm}, enabling them to think addresses a strong practical demand. Thinking enhances the reliability and performance of vision tasks~\citep{r1vl}, and task-specific SVLMs provide a compelling alternative to LVLMs in resource-constrained settings. This motivates the development of a new training paradigm that empowers SVLMs with thinking capabilities, at least for specialized tasks.



A promising solution is to fuse SFT and RLVR, as a well-calibrated \textbf{trade-off} can lower the high demands on the base model~\citep{deepseek_r1, yan2025learning}: SFT encourages the model to memorize verifiable thinking patterns to prevent advantage collapse, while RL forces exploration to prevent rigid templates from overwhelming the model's capacity. The central challenge, however, is that SVLMs struggle to achieve this balance. Existing hybrid methods, like two-stage training~\citep{chen2025sft, chu2025sftmrl} or annealed SFT losses\footnote{{See the supplementary material} for further comparison.}~\citep{zhang2025policy}, rely on a static trade-off governed by hyperparameters set empirically.
This rigidity is the critical flaw because the minimal capacity of SVLMs means the window for a successful static balance is incredibly narrow, making failure almost inevitable (\cf Fig. \ref{fig:fig1}b). Our repeated trials with two-stage training confirmed this issue, with performance often falling below the baseline (\cf Fig. \ref{fig:intro_demo}).

SVLMs therefore require a more intelligent paradigm to navigate this trade-off. To this end, we propose \model (\textbf{Dy}namic \textbf{M}emorize–\textbf{E}xplore), which integrates SFT and RLVR through a \textbf{dynamic} switching mechanism. As illustrated in Fig.~\ref{fig:fig1}b, \model assesses the model's generation at each step and adapts its training mode accordingly. When the model fails to follow instructions, it switches to a memorization mode (SFT) to guarantee stable optimization signals. Conversely, for valid generations, it engages an exploration mode (RLVR) to encourage diverse and grounded thinking. This state-driven approach ensures memorization and exploration are always complementary, dynamically maintaining the delicate trade-off. {While this dynamic switching alone guarantees training stability, we further maximize the model's potential by incorporating a synergistic \textit{Visual Supervision} mechanism. This module facilitates an adaptive interaction: the CoT ground-truth guides the scoring of exploration (via a visual checker), while successful exploration traces dynamically refine the CoT ground-truth (via a visual refiner).}




The aforementioned design makes \model a highly effective paradigm for empowering thinking in SVLMs for specific tasks. We validate this across three diverse domains, ranging from recognition-intensive tasks (medical VQA) to reasoning-intensive challenges (chart understanding and geometric problem solving). Remarkably, using only a few thousand training samples, \model achieves substantial performance gains, enabling it to match or even surpass several LVLMs. Our primary contributions are as follows:

\vspace{-0.5em}
\begin{enumerate}[leftmargin=*]
\itemsep-0.2em
\item We propose \model, the first training paradigm that equips SVLMs with thinking capabilities, substantially reducing reliance on the base VLM’s initial capacity.

\item Through dynamic switching and {synergistic} supervision, \model alleviates pseudo thinking traces and advantage collapse in SVLMs, yielding image-grounded thinking and consistent performance improvements.

\item We demonstrate the effectiveness and practicality of \model across three diverse domains, each consistently showing substantial performance gains with only a few thousand training samples.
\end{enumerate}

\section{Related Work}

\noindent\textbf{Vision-Language Models.}
Modern VLMs, such as LLaVA~\citep{liu2023improvedllava} and Qwen-VL~\citep{bai2023qwenvl}, have demonstrated remarkable capabilities across a wide array of vision tasks. 
However, their substantial parameter counts and computational demands restrict their use in resource-constrained environments like edge devices. This has motivated a growing interest in SVLMs designed for efficiency~\citep{zhou2024tinyllava, marafioti2025smolvlm, korrapati2024moondream}. Although works like TinyLLaVA~\citep{zhou2024tinyllava} and SmolVLM~\citep{marafioti2025smolvlm} have shown that carefully designed SVLMs can achieve competitive performance, they exhibit a critical weakness. Recent studies highlight that their performance degrades significantly on tasks requiring complex, multi-step instruction following, indicating a gap in their compositional understanding and general reasoning abilities~\citep{albalak2022data, ghosh2024exploring, liu2025phd}.

\noindent\textbf{Empowering Thinking Capabilities in VLMs.} 
Recent advances in LLM thinking (\eg, GPT-o1~\citep{openai_o1}, DeepSeek-R1~\citep{guo2025deepseek}) have motivated efforts to equip VLMs with similar capabilities via dedicated training paradigms. 

\underline{\emph{SFT on CoT data}}~\citep{xu2024llavacot, xia2024chartx, xia2024geox, gao2023gllava, yang2025r1one}.
This paradigm leverages large-scale CoT supervision to teach models to memorize and generalize thinking patterns. Multimodal-CoT~\citep{zhang2023multimodal} was an early attempt using fused visual–text inputs, but its small scale data limited genuine thinking. Subsequent works highlight the role of scale: G-LLaVA~\citep{gao2023gllava} constructs 170K geometry-specific CoT samples; ChartVLM~\citep{xia2024chartx} compiles a large chart corpus; and LLaVA-CoT~\citep{xu2024llavacot} as well as R1-OneVision~\citep{yang2025r1one} curate diverse, structured CoT data through large-scale prompt engineering. These approaches face long inputs, requiring large VLMs that can process rich textual information while preserving visual grounding~\citep{marafioti2025smolvlm, zhai2023investigating}.

\underline{\emph{RL with Verifiable Reward (RLVR)}}~\citep{r1vl, chen2025r1v, lmmr1, vlmr1, vrft}.
RLVR adopts a distinct paradigm that elicits thinking through autonomous exploration with minimal external supervision. The popularly used algorithm is Group Relative Policy Optimization (GRPO), introduced by DeepSeek-Math~\citep{shao2024deepseekmath}, which exploits models’ ability to produce structured outputs that separate thinking from final answers. It leverages rule-verifiable data to optimize high-scoring generations, while light SFT is employed for cold-start when the output structure is unclear. This paradigm has been extended to VLMs in several works. R1-V~\citep{chen2025r1v} applies GRPO to VLMs, enabling thinking in tasks such as counting and geometry. LMM-R1~\citep{lmmr1} introduces a two-stage pipeline that transfers textual thinking into multimodal learning. VisualRFT~\citep{vrft} and R1-VL~\citep{r1vl} incorporate vision-specific rewards to guide fine-grained, visually grounded optimization. Since GRPO depends on models’ initial structured thinking ability, these methods typically build on strong VLMs, such as the Qwen-VL series~\citep{Qwen2.5-VL}.

\underline{\emph{Hybrid Training Paradigms}}~\citep{chu2025sftmrl, yan2025learning, zhang2025policy}. To harness the complementary strengths of SFT and RL, researchers have also investigated hybrid paradigms. A common approach is a two-stage training process~\citep{chu2025sftmrl} that first uses SFT to teach the model the desired output format, followed by RL for exploration. Although intuitive, this method is highly sensitive to the amount of SFT, a parameter that is particularly challenging to tune for SVLMs, as these smaller models can easily become trapped in suboptimal states. Alternative strategies attempt to continuously blend SFT with RL, for instance, by incorporating SFT as an annealed auxiliary loss~\citep{zhang2025policy} or by managing its influence with an empirical shaping function~\citep{yan2025learning}. However, all these strategies ultimately rely on an empirically determined balance between the two paradigms. This rigidity represents a critical flaw when applied to SVLMs. The absence of adaptive control over the SFT weight renders these methods brittle and unreliable.

Thus, existing paradigms are not directly transferable to SVLMs due to their inherent limitations in model capacity and instruction-following ability. This highlights the need for a novel training paradigm that imposes minimal requirements on the base VLM.

\section{Approach}

\subsection{Preliminaries}

We first briefly recap the two training paradigms (SFT and RLVR) that underlie our method. Let $\mathcal{D}=\{(x_i,y_i)\}_{i=1}^{N}$ be the training set, where $x$ denotes the input (\eg an image-instruction pair) and $y$ the desired output.  The model defines a conditional distribution $p_\theta(y\mid x)$ with parameters~$\theta$.

\textbf{Supervised Fine-Tuning (SFT).}
For each training pair $(x,y)$ in $\mathcal{D}$, SFT updates the model by minimizing the negative log-likelihood (cross-entropy) of the desired output $y$ under the conditional distribution $p_\theta(y\mid x)$:
\begin{equation}
    \mathcal{L}_{\text{SFT}}(\theta)
 = -\mathbb{E}_{(x,y)\sim\mathcal{D}}
   \bigl[\log p_\theta(y\mid x)\bigr].
   \label{eq:sft}
\end{equation}

This teacher-forcing loss allows models to \emph{memorize} extensive training examples, compelling the model to absorb this knowledge.

\textbf{Group Relative Policy Optimization (GRPO).} GRPO is an RL algorithm that \emph{explores} open-ended generation by comparing candidate outputs within a group.  For each input $x$, the policy $p_\theta$ samples a set $\{\tilde{y}^k\}_{k=1}^{K}$; a reward function $r_a(\tilde{y}^k)$ is computed based on the correctness of the output answer, and each sample’s advantage $A$ is measured relative to the other group members:
\begin{equation}
\small
A(\tilde{y}^k)\;=\;
\frac{r_a(\tilde{y}^k)\;-\;\bar{r}_a}
     {\sigma+\varepsilon},\quad
\bar{r}_a=\tfrac{1}{K}\sum_{j=1}^{K}r_a(\tilde{y}^j),\; \quad
\sigma=\sqrt{\tfrac{1}{K}\sum_{j=1}^{K}(r_a(\tilde{y}^j)-\bar{r})^2},
\label{eq:adv}
\end{equation}
where $\varepsilon$ is a small constant for numerical stability. The policy then updates its parameters by minimizing the following loss, regularised by a KL constraint:
\begin{align}
\small
\mathcal{L}_{\text{GRPO}}(\theta)
  &=
  -\mathbb{E}_{x\sim\mathcal{D}}\,\mathbb{E}_{\tilde{y}\sim p_\theta}
    \Bigl[
      \min\bigl(
        r_{\theta}(x, \tilde{y})\,A(\tilde{y}),
        \operatorname{clip}\bigl(r_{\theta}(x, \tilde{y}); 1-\epsilon,1+\epsilon\bigr)
        A(\tilde{y})
      \bigr)
    \Bigr] \notag\\[4pt]
  &\quad
  +\beta\,D_{\mathrm{KL}}\bigl[
        p_\theta(\cdot\!\mid\!x)\,\|\,p_{\text{ref}}(\cdot\!\mid\!x)
     \bigr], \quad \text{where} \quad r_{\theta}(x, \tilde{y})\;=\;
\frac{p_\theta(\tilde{y}\mid x)}{p_{\text{old}}(\tilde{y}\mid x)}.
     \label{eq:grpo-loss}
\end{align}
%
The clip and KL terms work together to keep each update close to safe regions of the parameter space: the clip gate limits step size around the rollout policy $p_{\text{old}}$, while the KL term ($\beta D_{\mathrm{KL}}$) tethers the policy to the reference $p_{\text{ref}}$ (typically the initial model).


\textbf{Gradient Compatibility of SFT and GRPO.}
Below, we reveal that the optimization objectives of SFT and GRPO are formally equivalent, with the former targeting the ground-truth data distribution and the latter an internal one.


The gradient of the SFT loss is straightforward:
\begin{equation}
\nabla_\theta \mathcal{L}_{\mathrm{SFT}}(\theta)
= -\mathbb{E}_{(x,y)\sim\mathcal{D}}
\left[\nabla_\theta \log p_\theta(y\mid x)\right].
\end{equation}

Similarly, the GRPO gradient (ignoring clipping and any KL‐penalty) can be written as
\begin{equation}
\label{eq:grpo-gradient}
\nabla_\theta \mathcal{L}_{\rm GRPO}(\theta)
= -\,\mathbb{E}_{\substack{x\sim\mathcal{D},\\\tilde y\sim p_{\rm old}(\cdot\mid x)}}
  \bigl[r_{\theta}(x, \tilde{y})\,A(\tilde y)\,\nabla_\theta\log p_\theta(\tilde y\mid x)\bigr].
\end{equation}

This comparison shows that the SFT gradient is a special case of the GRPO gradient, obtained when the ground-truth sample is used with unit advantage. This equivalence enables a unified loss that balances external imitation (SFT) with internal refinement (GRPO). Achieving this fusion requires dynamically weighting the two signals (\S \ref{sec:dyme}) and ensuring stylistic consistency between external ground-truth and self-generated outputs (\S\ref{sec:vm}).


\subsection{Dynamic Memorize–Explore (\model)}
\label{sec:dyme}
To realize this complementarity, we propose the \textbf{Dy}namic \textbf{M}emorize–\textbf{E}xplore (\model) paradigm, which adaptively switches between SFT and GRPO at each training step. In the following, we first outline the overall pipeline and then elaborate on the optimization procedures for each mode.

\begin{figure*}[!t]
  \centering
  \begin{subfigure}[b]{0.6\textwidth}  
    \centering
    \includegraphics[
      clip,
      trim=1cm 0cm 15cm 0.cm,  
      height=6.3cm
    ]{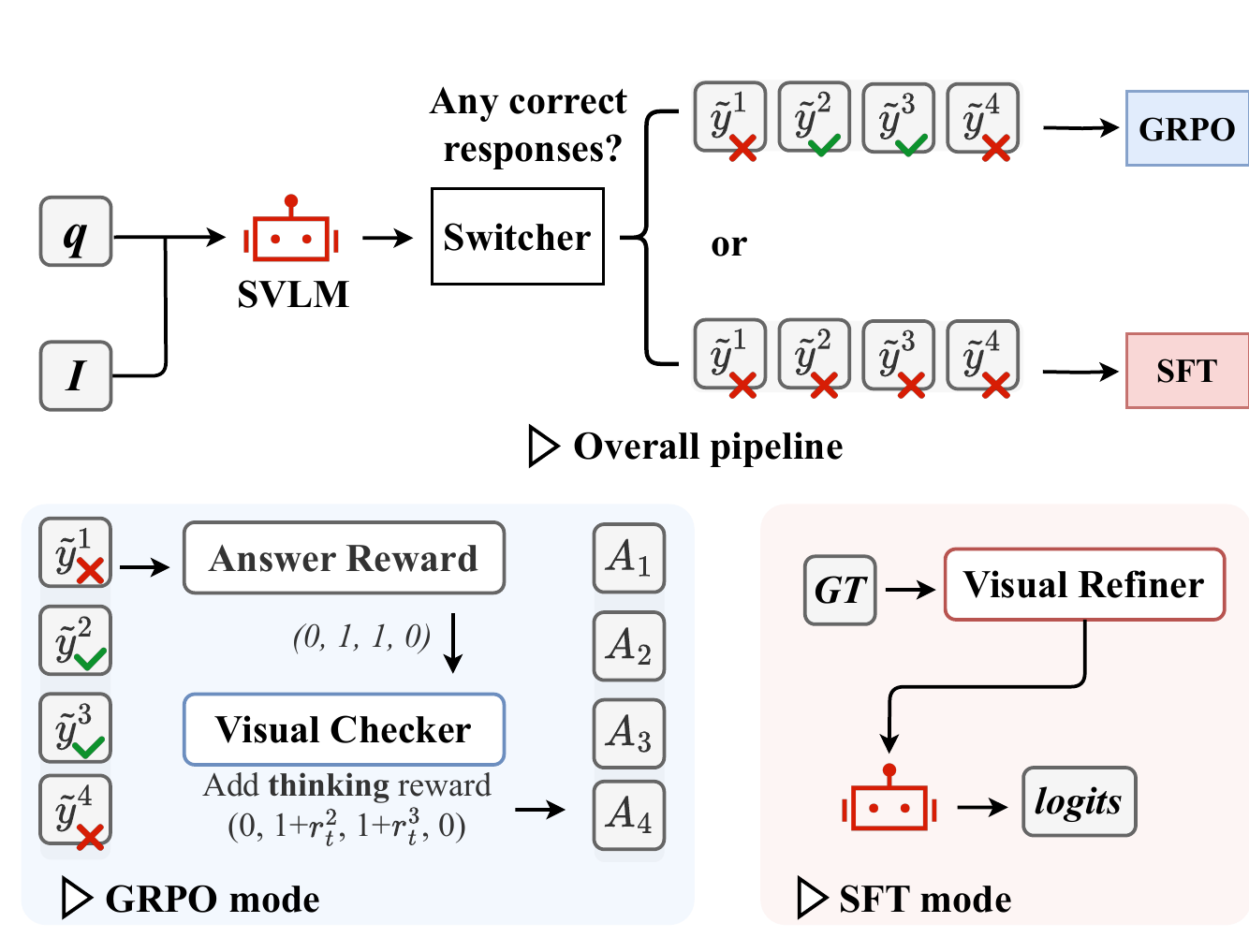}
    \caption{The pipeline for \texttt{\model}.}
    \label{fig:left-part}
  \end{subfigure}%
  \begin{subfigure}[b]{0.36\textwidth}  
    \centering
    \includegraphics[
      clip,
      trim=24.8cm 0cm 0cm 0.cm,  
      height=6.3cm
    ]{Template/figures/pipeline.pdf}
    \caption{Visual refiner and checker.}
    \label{fig:right-part}
  \end{subfigure}
  \caption{\textbf{Workflow and module components of \model.} At each training step, \model dynamically switches between memorization (via SFT) and exploration (via GRPO) modes based on its generations. Visual supervision is introduced through the visual refiner and visual checker. The refiner enhances the targets for memorization by incorporating richer visual elements (green), while the checker rewards the thinking context generated based on their visual relevance.}
  \label{fig:dyme-pipeline}
\end{figure*}

\textbf{Overall.} As shown in Fig.~\ref{fig:left-part}, each training step begins with an input $x = (I, q)$, where $I$ is the image and $q$ is an instruction. The policy SVLM $p_\theta$ generates $K$ responses $\{\tilde{y}^k\}_{k=1}^K$. Each response is parsed into a thinking trace and a final answer, which is then verified for correctness using predefined rules. The verification results fall into two categories: either all responses are incorrect  (including those that fail to parse), or at least one is correct.
\textbf{The decision rule:} if at least one response is correct, the model proceeds with GRPO-based exploration; otherwise, it falls back to SFT-based memorization. Formally, the training mode is switched as:
\begin{equation}
\text{mode}(x) =
\begin{cases}
\text{GRPO}, & \text{if } \max_k r_a(\tilde{y}^k) = 1, \\
\text{SFT},  & \text{otherwise},
\end{cases}
\end{equation}
where $r_a(\tilde{y}^k) \in \{0, 1\}$ indicates whether $\tilde{y}^k$ passes rule-based verification.
Though simple, this decision rule is highly effective. When all responses are incorrect, the answer rewards are essentially all zero and the normalized advantages become dominated by noise, making GRPO updates for a small SVLM unstable. In this regime, falling back to SFT provides a low-variance, ground-truth guided gradient. Conversely, the appearance of at least one correct response indicates that the current policy has already discovered a feasible solution for this input, so GRPO can safely exploit the relative advantages to drive exploration.


\textbf{GRPO Mode.} \model introduces a key refinement to the original GRPO: beyond the answer reward $r_a$, it incorporates an auxiliary reward $r_t$ for thinking traces. {This reward is computed by evaluating the generated traces against expected thinking patterns (\eg, via token-level F1 score ground-truth comparison), promoting structured thinking.}


Given these rewards, we update the policy using a modified GRPO objective. Unlike the standard formulation (Eqs.~\ref{eq:adv} \& \ref{eq:grpo-loss}), we omit the KL penalty and clipping terms, as the dynamic integration of SFT already stabilizes training. This avoids overly conservative updates and yields a cleaner gradient form, enabling smoother alignment between SFT and GRPO:
\begin{equation}
\label{eq:dyme-grpo-loss}
\tilde{\mathcal{L}}_{\text{GRPO}}(\theta)
= -\mathbb{E}_{x\sim\mathcal{D}}\,\mathbb{E}_{\tilde{y} \sim p_\theta(\cdot\mid x)}
  \left[ r_\theta(x, \tilde{y})\,  A(\tilde{y})\right],
\end{equation}
where $A(\tilde{y}^k)$ is the group-normalized advantage calculated from the combined answer ($r_a$) and thinking ($r_t$) rewards, and $r_\theta(x, \tilde{y}^k) = \frac{p_\theta(\tilde{y}\mid x)}{p_{\text{old}}(\tilde{y}\mid x)}$ is the importance sampling ratio.

\textbf{SFT Mode.} When training falls back to SFT, the model is optimized toward the ground-truth response $y$ using the standard supervised loss (Eq.~\ref{eq:sft}). {This ensures that whenever the model fails to explore effectively, it receives a stable, ground-truth-guided gradient update to correct its behavior.}



\textbf{\model Objective.} The final loss dynamically combines the two objectives based on response correctness:
\begin{equation}
\mathcal{L}_{\text{DyME}}(\theta) =
\mathbbm{1}\left[ \max_k r_a(\tilde{y}^k) = 1 \right] \cdot \tilde{\mathcal{L}}_{\text{GRPO}}(\theta)
+
\left(1 - \mathbbm{1}\left[ \max_k r_a(\tilde{y}^k) = 1 \right]\right) \cdot \mathcal{L}_{\text{SFT}}(\theta),
\end{equation}
\begin{flushleft}
where $\mathbbm{1}[\cdot]$ is the indicator function, returning 1 if the condition holds, 0 otherwise.
\end{flushleft}

\subsection{Vision Supervision}
\label{sec:vm}

{\textbf{\model with Visual Supervision.} While the aforementioned \textbf{Pure \model} (using standard $r_t$ and static ground-truth) already guarantees training stability through its dynamic switching mechanism, we can further exploit this dynamic nature to maximize performance. Specifically, the switching mechanism allows us to tailor the supervision signals at each optimization step: refining the reward during exploration and enhancing the ground-truth during memorization. To this end, we introduce a \emph{checker--refiner} framework (\cf Fig.~\ref{fig:right-part}), which constitutes the \textbf{Full \model}.}

This framework reorganizes the ground-truth to adhere to a predefined structure, crucially transforming it into a grounded thinking trace. The refiner restructures the external ground-truth into structured, visually grounded responses, while the checker evaluates self-generated outputs for their structural organization and coverage of visual content. We refer to the resulting supervision signals collectively as \emph{vision supervision}. The implementation is carried out via LLM-based prompt engineering.

\textbf{Visual Facts $I_c$} are central to realizing vision supervision. They are defined as fine-grained visual components extracted from each image, including objects, attributes, and states. These elements play a dual role: they provide evidence for evaluating generations against the image and serve as building blocks for constructing complete ground-truth responses.

\textbf{Visual Checker.} The visual checker evaluates responses along two dimensions: (i) whether the output contains sufficient correct visual elements compared to $I_c$, and (ii) whether it aligns stylistically with provided examples. 
These examples may be manually defined or extracted from the SFT ground-truth.

\textbf{Visual Refiner.} 
The refiner produces visually grounded responses for SFT by leveraging the model’s validated explorations. High-scoring traces identified by the visual checker are stored in a dynamic example pool. An LLM then draws from this pool to generate ground-truth responses, integrating structural templates with visual facts from $I_c$ and referencing the collected examples.

{In essence, the acquisition of Visual Facts, the evaluation by the Visual Checker, and the synthesis by the Visual Refiner are all implemented via structured prompt engineering using Qwen2.5-14B. Please refer to the Supplementary Materials for the full prompts used in our pipeline.}


\section{Experiments}
{To rigorously evaluate \model, we structure our experiments into two parts: (1) \textbf{Algorithmic Validation}, where we evaluate ``Pure \model'' in a controlled setting using offline data to isolate the contribution of our dynamic switching mechanism; and (2) \textbf{System Effectiveness}, where we evaluate the full \model pipeline (with Visual Supervision) across diverse domains to demonstrate its practical capability in empowering SVLMs.

\subsection{Part I: Algorithmic Validation (Pure \model)}
\label{sec:pure_dyme}

\textbf{Setup.} Since SVLMs lack intrinsic reasoning capabilities and cannot autonomously discover complex reasoning paths, pre-constructed CoT data is a mandatory prerequisite for all training paradigms. 
We therefore evaluated all methods on ChartQA~\citep{masry-etal-2022-chartqa} using LLaVA-OV-S~\citep{llavaov}, the 0.5B variant, with three pre-constructed CoT datasets of varying qualities: \textbf{Low (Undesigned)} containing unstructured traces ($\sim$80 words); \textbf{Medium (Standard)} consisting of semi-structured traces ($\sim$89 words) from Qwen2.5-14B; and \textbf{High (Premium)} comprising highly structured traces ($\sim$142 words) from GPT-4o. Following established protocols~\citep{liu-etal-2023-deplot,masry-etal-2022-chartqa}, we report \emph{relaxed correctness}, which allows a 5\% tolerance for numerical answers.

We present a threefold evaluation to validate data robustness, design optimality, and generalization:

\textbf{(1) Robustness to Data Quality.} Table~\ref{tab:pure_dyme_analysis}~(a) demonstrated \model's superiority. On \emph{Low} quality data, Pure \model ($61.9\%$) significantly outperforms the unstable Two-stage baseline ($57.6\%$). Remarkably, using only \emph{Medium} data, it surpasses the SFT baseline trained on premium \emph{High} (GPT-4o) data ($61.6\%$). This confirms that \model acts as a robust student, effectively maximizing data efficiency.

\textbf{(2) Optimality of Binary Switching.} To validate our binary design, we compared it against three alternative switching heuristics in Table~\ref{tab:pure_dyme_analysis}~(b): 
(i) \textit{Reward Thresholding}, which switches to RL only if the batch average reward exceeds a threshold $t$; 
(ii) \textit{SFT Annealing}, which applies a weighted SFT loss alongside RL at every step; and 
(iii) \textit{SFT Budget}, which performs focused SFT updates on accumulated failure cases (hard mining).

\textbf{Results:} \textit{Reward Thresholding} proves brittle, collapsing at suboptimal thresholds ($t=0.5$, $52.4\%$). \textit{SFT Annealing} incurs a heavy computational tax ($+25\%$) due to the auxiliary SFT gradient calculation. \textit{SFT Budget} yields inferior results ($59.6\%$) as overwhelming the model with concentrated failures destabilizes learning. In contrast, \model's binary switch is parameter-free, efficient, and empirically optimal ($64.9\%$).

\begin{table}[h]
\centering
\caption{{\textbf{Algorithmic Validation of Pure \model.} (a) \model outperforms SFT and Two-stage variants (w/ and w/o KL penalty) across all data qualities. (b) The binary switch is more robust and efficient than soft or hard-mining alternatives (evaluated on Medium data).}}
\label{tab:pure_dyme_analysis}
\resizebox{\linewidth}{!}{
\begin{tabular}{@{}lccc|llcc@{}}
\toprule
\multicolumn{4}{c|}{\textbf{(a) Robustness across Data Quality}} & \multicolumn{4}{c}{\textbf{(b) Switching Strategy Ablation}} \\
\cmidrule(r){1-4} \cmidrule(l){5-8}
\textbf{Method} & \textbf{Low} & \textbf{Medium} & \textbf{High} & \textbf{Strategy} & \textbf{Hyperparam.} & \textbf{Acc.} & \textbf{Cost} \\
\midrule
SFT & 50.5 & 57.8 & 61.6 & Reward Threshold & $t=0.5/0.8/0.9$ & 52.4/64.1/63.4 & None \\
Two-stage & 57.6 & 59.9 & 54.5 & SFT Annealing & Cosine & 64.0 & +25\% \\
Two-stage (w/ KL) & 55.4 & 60.8 & 62.7 & SFT Budget & Hard Mining & 59.6 & Budget-dep. \\
\rowcolor{gray!10} \textbf{Pure \model} & \textbf{61.9} & \textbf{64.9} & \textbf{68.5} & \textbf{Binary Switch (Ours)} & \textbf{--} & \textbf{64.9} & \textbf{Baseline} \\
\bottomrule
\end{tabular}
}
\end{table}

\textbf{(3) Mechanism Generality.} Going beyond the primary setup, while \model is primarily tailored for SVLMs, we verify the universality of its core switching mechanism (see Supplementary). 
In the text-only domain, it boosts the small-scale Qwen2.5-0.5B on GSM8K~\citep{gsm8k} to $55.3\%$ (+5.8\% over GRPO), confirming \model is an effective paradigm for empowering thinking in small-parameter models regardless of modality. 
Moreover, the paradigm scales effectively: on the stronger Qwen2.5-VL-7B, it further improves ChartQA performance to $89.6\%$ (+2.3\%).
}

{\subsection{Part II: System Effectiveness (Full \model)}}
Having validated the algorithmic core, we now evaluate the Full \model pipeline, augmented with Visual Supervision, across three diverse domains: Medical VQA, Chart Understanding, and Geometry. Each followed the evaluation protocols of prior work~\citep{mova}.

\textbf{Setup \& Source of $I_c$.} Unlike Part I, here we activate the Visual Supervision module to enable the full online loop. {Crucially, to demonstrate \model's capability to bootstrap from raw signals, we utilize the ``Undesigned'' CoT data (defined in \S\ref{sec:pure_dyme}) derived from SLAKE~\citep{liu2021slake}, ChartQA~\citep{masry-etal-2022-chartqa}, and Geo170K~\citep{gao2023gllava} as the common training source for all methods.}
Acquiring the necessary visual facts ($I_c$) is a fully automated process: we leverage standard domain tools (\eg, BiomedGPT~\citep{BiomedGPT} for medical, DePlot~\citep{liu-etal-2023-deplot} for charts) or prompt generalist LLMs (\eg, Qwen2.5~\citep{qwen2.5}) to parse images into structured textual descriptions. {The automated pipeline and prompts are included in the supplementary.
}

{\textbf{Evaluation Protocol.} We used official train-test splits for SLAKE (Accuracy/Recall) and ChartQA (Relaxed correctness). For Geometry, since Geo170K~\citep{gao2023gllava} provides no test set, we evaluated Accuracy on MathVerse~\citep{zhang2024mathverse}, consistent with \cite{mova}.

}

\subsubsection{Main Results}

\begin{wrapfigure}{r}{0.4\linewidth}
    \centering
    \includegraphics[width=\linewidth]{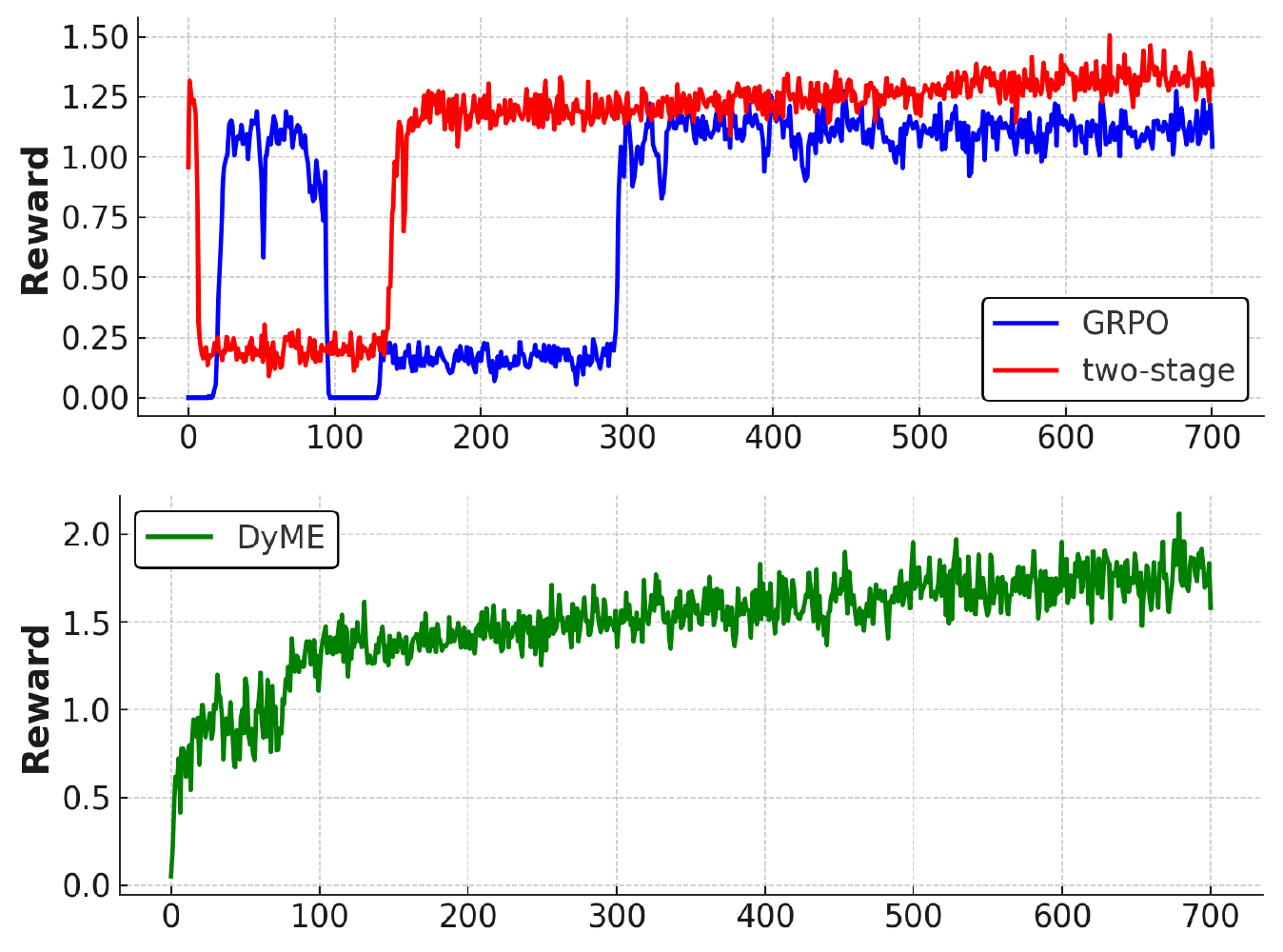}
    \caption{\textbf{Training rewards.} GRPO and two-stage training suffer from severe advantage collapse.}
    \label{fig:stable}
\end{wrapfigure}

\textbf{\model \mvs{ Existing Training Paradigms}}.
The comprehensive results in Table~\ref{tab:three_domains} show that \model consistently delivers substantial gains. Notably, after training with \model, SmolVLM improves from $49.9$ to $55.6$ (+$5.7$), LLaVA-OV-S from $50.7$ to $55.4$ (+$4.7$), and InternVL2-S from $56.3$ to $58.1$ (+$1.8$). 
In contrast, existing paradigms tend to degrade performance (\eg, SFT lowers SmolVLM to $44.1$), validating our analysis that SFT yields pseudo thinking traces and GRPO faces advantage collapse (\cf Fig.~\ref{fig:stable}).

\begin{figure}[t]
    \centering
    \includegraphics[width=.95\linewidth]{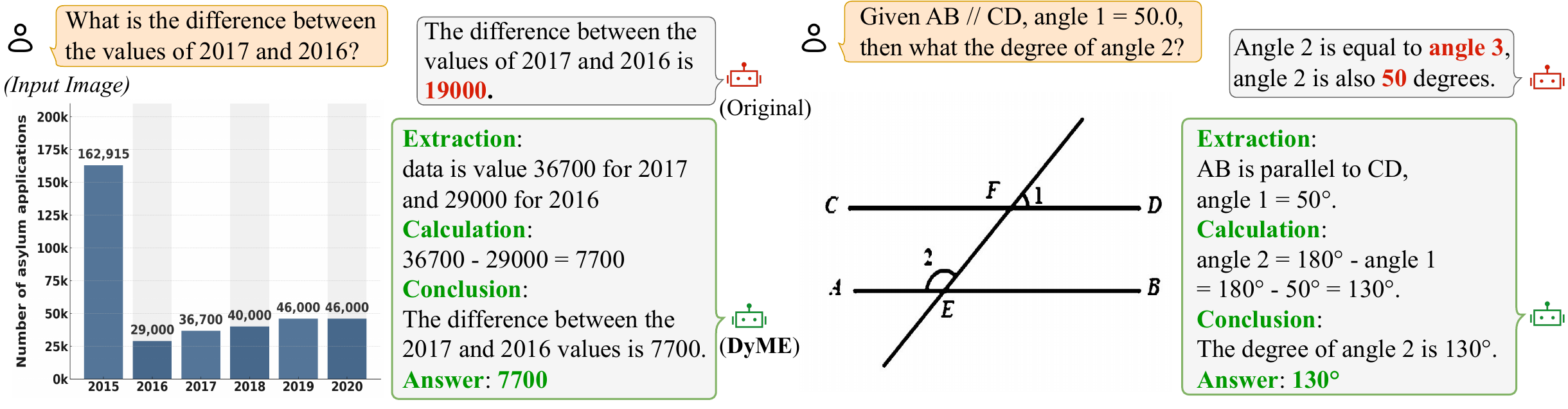}
    \caption{\textbf{Showcases on chart understanding and geometry solving.} We use LLaVA-OV-S to demonstrate the results. The SVLM originally produces hallucinated answers (red), while the DyME-trained model generates structured thinking traces (green) that incorporate grounded values, effectively improving the performance.}
    \label{fig:showcases}
\end{figure}

\model effectively mitigates these issues. It promotes grounded traces that are concise yet informative (\cf Fig.~\ref{fig:showcases}), aligning well with the limited capacity of SVLMs. Importantly, \model places minimal demands on the base model: even SmolVLM (0.5B) achieves substantial gains, and it still delivers improvements (+$2.6$\%) on extensively pretrained models like InternVL2-S. {We further corroborated these findings through manual inspection, as detailed in the Supplementary Material.}

{\textbf{Matching the Efficacy of GPT-4o Supervision with Open-Source Models.} 
Comparing results between Part I and Part II reveals a crucial finding: LLaVA-OV-S trained with the full \model pipeline (using the accessible Qwen2.5-14B) achieves {67.5\%} (Table~\ref{tab:three_domains}). This effectively matches the performance of Pure \model trained on expensive {GPT-4o} data ({68.5\%}, \cf Table~\ref{tab:pure_dyme_analysis}). 
This proves that full \model allows open-source supervision to achieve training outcomes comparable to those derived from top-tier proprietary models, eliminating the need for expensive data annotation.}

\begin{table}[t]
    \centering
    \caption{\textbf{Comparisons across three domains: medical VQA, chart understanding, and geometry solving.} The evaluation follows the VLMEvalKit framework~\citep{duan2024vlmevalkit}. For SVLMs, existing training paradigms degrade their performance, whereas \model consistently brings improvements. The best performance achieved by each SVLM is highlighted in bold, with the relative improvement also indicated. Notably, after being trained with \model, SVLMs achieve performance comparable to that of MoVA (underlined).}
    \label{tab:three_domains}
    
    \vspace{2pt}

    \resizebox{0.9\linewidth}{!}{
        \footnotesize
        \setlength{\tabcolsep}{3pt} 
        
        \begin{tabularx}{\textwidth}{
            l                              
            l                              
            l                              
            >{\raggedleft\arraybackslash}X  
            >{\raggedleft\arraybackslash}X  
            >{\raggedleft\arraybackslash}X  
            >{\raggedleft\arraybackslash}X  
        }
        \toprule
        \textbf{Model} 
          & \textbf{ViT} 
          & \textbf{LLM} 
          & \textbf{Medical} 
          & \textbf{Chart}
          & \textbf{Geometry}
          & \textbf{Avg.} \\
        \midrule
        \multicolumn{7}{l}{\textbf{LVLMs}} \\[0.2em]
        LLaVA-Med~\citep{li2023llavamed}      & CLIP-ViT-L/14  & Vicuna-7B        & 64.3  & --   & --   & -- \\ 
        Cambrian-1~\citep{tong2024cambrian}     & Hybrid-3B        & Llama3-8B        & --    & 72.6 & 22.0   & -- \\ 
        LLaVA-1.5~\citep{liu2023improvedllava}      & CLIP-ViT-L/14  & Vicuna-7B        & 69.4  & 17.8 & --  & -- \\ 
        LLaVA-1.6~\citep{liu2024llavanext}      & CLIP-ViT-L/14  & Vicuna-7B        & 78.2    & 49.2 & 13.4 & 47.0 \\ 
        MoVA~\citep{mova}        & Hybrid-3B  & Vicuna-7B        & 74.5  & 68.3 & 19.7 & \underline{54.2} \\ 
        LLaVA-OV-L~\citep{llavaov}     & SigLIP-SO400M  & Qwen2-7B         & 75.7     & 80.9 & 24.5 & 60.4 \\ 
        InternVL2-L~\citep{intern}     & InternViT-300M  & InternLM2.5-7B        & 80.2     & 82.1 & 37.3 & 66.5 \\ 
        
        \addlinespace
        \multicolumn{7}{l}{\textbf{SVLMs}} \\[0.2em]
        SmolVLM~\citep{marafioti2025smolvlm}        & SigLIP-93M     & SmolLM2-360M     & 72.1  & 63.2 & 14.6 & 49.9 \\ 
        \quad + CoT SFT & SigLIP-93M    & SmolLM2-360M     & 60.1  & 57.7 & 14.5 & 44.1 \\ 
        \quad + GRPO    & SigLIP-93M    & SmolLM2-360M     & 61.1  & 53.8 & 17.1 & 44.0 \\ 
        \quad + Two-stage & SigLIP-93M  & SmolLM2-360M     & 59.4  & 60.1 & 16.7 & 45.4 \\ 
        
        \rowcolor{gray!30}
        \quad \textbf{+ \model}
          & SigLIP-93M
          & SmolLM2-360M
          & {\cellcolor{gray!30}\raggedleft
              \textbf{78.1}\\[-0.5ex]\textbf{\scriptsize(+6.0\%)}}  
          & {\cellcolor{gray!30}\raggedleft
              \textbf{69.7}\\[-0.5ex]\textbf{\scriptsize(+6.5\%)}}  
          & {\cellcolor{gray!30}\raggedleft
              \textbf{18.9}\\[-0.5ex]\textbf{\scriptsize(+4.3\%)}}  
          & {\cellcolor{gray!30}\raggedleft
              \textbf{55.6}\\[-0.5ex]\textbf{\scriptsize(+5.7\%)}}  
          \\
        
        \addlinespace
        LLaVA-OV-S~\citep{llavaov}     & SigLIP-400M  & Qwen2-0.5B       & 74.9  & 61.4 & 15.9 & 50.7 \\ 
        \quad + Two-stage & SigLIP-400M & Qwen2-0.5B    & 74.5  & 52.9 & 16.5 & 48.0 \\  
        \rowcolor{gray!30}
        \quad \textbf{+ \model}
          & SigLIP-400M
          & Qwen2-0.5B
          & {\cellcolor{gray!30}\raggedleft
              \textbf{78.3}\\[-0.5ex]
              \textbf{\scriptsize(+3.4\%)}}
          & {\cellcolor{gray!30}\raggedleft
              \textbf{67.5}\\[-0.5ex]
              \textbf{\scriptsize(+6.1\%)}}
          & {\cellcolor{gray!30}\raggedleft
              \textbf{20.4}\\[-0.5ex]
              \textbf{\scriptsize(+4.5\%)}}
          & {\cellcolor{gray!30}\raggedleft
              \textbf{55.4}\\[-0.5ex]
              \textbf{\scriptsize(+4.7\%)}}
          \\
        
        \addlinespace
        InternVL2-S~\citep{intern}    & InternViT-300M & Qwen2-0.5B       & 78.3  & 71.9 & 18.7   & 56.3 \\ 
        \quad + Two-stage 
          & InternViT-300M
          & Qwen2-0.5B
          & 73.6
          & 55.7
          & 17.1
          & 48.8
          \\
        \rowcolor{gray!30}
        \quad \textbf{+ \model}
          & InternViT-300M
          & Qwen2-0.5B
          & {\cellcolor{gray!30}\raggedleft
              \textbf{80.0}\\[-0.5ex]\textbf{\scriptsize(+1.7\%)}}  
          & {\cellcolor{gray!30}\raggedleft
              \textbf{74.5}\\[-0.5ex]\textbf{\scriptsize(+2.6\%)}}  
          & {\cellcolor{gray!30}\raggedleft
              \textbf{19.8}\\[-0.5ex]\textbf{\scriptsize(+1.1\%)}}  
          & {\cellcolor{gray!30}\raggedleft
              \textbf{58.1}\\[-0.5ex]\textbf{\scriptsize(+1.8\%)}}  
          \\
        
        \bottomrule
        \end{tabularx}
    }
\end{table}


\textbf{\model-trained SVLMs Can Be Competitive with LVLMs.}
We ensured fairness by exposing all baselines to our training data. As shown in Table~\ref{tab:three_domains}, SVLMs trained with \model can surpass stronger LVLMs like MoVA ($54.2$) on these specialized domains, with SmolVLM reaching $55.6$ and LLaVA-OV-S $55.4$. As a result, \model-trained SVLMs become reliable options for task-specific applications on resource-constrained edge devices.

\subsubsection{Ablation Study}

To dissect the source of these gains, we conducted an ablation study to analyze the contribution of \model's four core components within the full pipeline: the memorization mode, exploration mode, visual refiner, and visual checker. Table~\ref{tab:ablation-dyme} shows the performance impact.

\begin{wraptable}{r}{0.5\linewidth}
\centering
\captionof{table}{\textbf{Ablation study.} Model: LLaVA-OV-S.}
\label{tab:ablation-dyme}
\resizebox{\linewidth}{!}{
\begin{tabular}{@{}lrrrr@{}}
\toprule
\textbf{\model Variant} & \textbf{Medical} & \textbf{Chart} & \textbf{Geometry} & \textbf{Average} \\
\midrule
\model (full) & 78.3 & 67.5 & 20.4 & 55.4 \\
\quad w/o memorization & 63.2 & 53.4 & 15.0 & 43.9 \drop{20.6} \\
\quad w/o exploration & 75.5 & 61.3 & 14.5 & 50.4 \drop{9.0} \\
\quad w/o visual refiner & 75.6 & 62.3 & 16.8 & 51.6 \drop{6.9} \\
\quad w/o visual checker & 76.9 & 64.3 & 17.1 & 52.8 \drop{4.7} \\
\bottomrule
\end{tabular}
}
\end{wraptable}


\textbf{Dynamic Switching Mechanism.} The results confirm that Memorization and Exploration are symbiotic. Disabling memorization causes a catastrophic drop ($55.4 \to 43.9$), effectively reverting to unconstrained, unstable exploration. Conversely, removing exploration ($50.4$) restricts the model to the static imitation of suboptimal data. As shown in Fig.~\ref{fig:stable}, their dynamic interplay prevents the advantage collapse observed in baselines, ensuring optimization stability throughout the learning process.

{\textbf{Visual Supervision.} Removing the visual checker and refiner drops performance by $4.7\%$ and $6.9\%$, respectively. This validates the pivotal role of visual supervision in bootstrapping from noisy, undesigned data. Given the limited capacity of SVLMs, they are easily prone to hallucination when trained on low-quality traces. The visual components act as a dynamic denoiser, ensuring that raw, imperfect data is filtered and refined into grounded visual facts ($I_c$) before optimization, thus enabling robust learning even from weak supervision.}

{\subsection{Training Efficiency \& Discussion}
\label{sec:efficiency}

We analyze the computational efficiency and performance trade-offs associated with different configurations of \model. The comparative results are detailed in Table~\ref{tab:efficiency}.
}

\begin{wraptable}{r}{0.4\linewidth} 
    \raa{0.9}
    \centering
    \captionof{table}{{\textbf{Cost-Benefit Analysis.} Time measured in sec/step. Run on 8x H800.}}
    \label{tab:efficiency}
    \resizebox{\linewidth}{!}{
    \color{black}
        \begin{tabular}{llcc}
        \toprule
        \textbf{Method} & \textbf{Ext. Model} & \textbf{Time} & \textbf{Acc.} \\
        \midrule
        GRPO (Baseline) & Qwen2.5-14B$^\dagger$ & 14.8s & 60.8 \\
        Pure \model     & Qwen2.5-14B$^\dagger$      & 14.0s & 64.9 \\
        Pure \model     & GPT-4o$^\dagger$      & 19.1s & \textbf{68.5} \\
        \midrule
        {Full \model}     & Qwen2.5-7B   & 21.2s & 66.8 \\
        {Full \model}      & Qwen2.5-14B  & 23.4s & 67.5 \\
        \bottomrule
        \multicolumn{4}{l}{\footnotesize $^\dagger$ Used for offline data construction only.}
        \end{tabular}
    }
\end{wraptable}
{
\textbf{Computational Efficiency \mvs Data Cost.} 
The framework offers two distinct operating regimes catering to different resource profiles. 
{Pure \model} represents the high-efficiency regime: when offline CoT data is pre-constructed, it maintains training throughput comparable to standard GRPO ($\sim$14s/step) while delivering superior performance. 
In contrast, {Full \model} (with Visual Supervision) prioritizes data autonomy. While the online interaction introduces a computational overhead ($\sim$1.6$\times$ training time), it enables the model to bootstrap high-performance reasoning solely from open-source models, bypassing the dependency on expensive, proprietary data annotation (\eg, GPT-4o).

\textbf{Sensitivity to External Model Capacity.}  For Full \model, we further examine the impact of the external helper's size on system performance. As shown in Table~\ref{tab:efficiency}, replacing the Qwen2.5-14B helper with the smaller 7B variant results in a negligible performance variation ($67.5\% \to 66.8\%$). This indicates that our structured prompt engineering effectively decomposes complex reasoning tasks, allowing even smaller external models to provide sufficient guidance for SVLMs without necessitating heavy-weight models.

\textbf{Applicability of Visual Supervision.} 
The effectiveness of the Visual Supervision module relies on the explicit extraction of Visual Facts ($I_c$). This process creates specific applicability boundaries. For domains involving \emph{abstract semantics} (\eg, irony in memes) or \textit{unstructured perception} (\eg, dense crowds), converting holistic visual signals into discrete text may result in information loss. In such scenarios, reverting to the Pure \model paradigm serves as a more robust alternative.
}

\section{Conclusion}
In this work, we introduced \model, a novel training paradigm designed to empower thinking capabilities within SVLMs. At its core, \model combines memorization (via SFT) mode and exploration (via RLVR) mode through a dynamic switching mechanism. Our experiments demonstrate that this approach not only resolves the critical trade-off between these two modes but also yields substantial performance gains on a wide spectrum of vision tasks, from recognition-intensive to reasoning-intensive scenarios. The success of \model is attributed to its carefully designed components: the dynamic switching mechanism addresses pseudo thinking traces and advantage collapse, while the visual checker and refiner provide coordinated, high-quality visual supervision. It imposes minimal requirements on the base VLM, making it broadly applicable to a wide range of models, including extremely lightweight SVLMs. Therefore, \model serves as the practical solution for empowering SVLMs to think.



\section*{Acknowledgment}
This work was supported by the Hong Kong SAR RGC General Research Fund (16219025), National Natural Science Foundation of China Young Scholar Fund Category B (62522216), National Natural Science Foundation of China Young Scholar Fund Category C (62402408), and Hong Kong SAR RGC Early Career Scheme (26208924).

{
\bibliography{iclr2026_conference}

@String(CVPR= {IEEE Conf. Comput. Vis. Pattern Recog.})

@String(ECCV= {Eur. Conf. Comput. Vis.})

@String(ICLR = {Int. Conf. Learn. Represent.})

@String(CVPR  = {CVPR})

@String(ECCV  = {ECCV})

@String(ICLR  = {ICLR})

@article{vlmr1,
  title={{VLM-R1}: A stable and generalizable {R1}-style large vision-language model},
  author={Shen, Haozhan and Liu, Peng and Li, Jingcheng and Fang, Chunxin and Ma, Yibo and Liao, Jiajia and Shen, Qiaoli and Zhang, Zilun and Zhao, Kangjia and Zhang, Qianqian and Xu, Ruochen and Zhao, Tiancheng },
  journal={arXiv preprint arXiv:2504.07615},
  year={2025}
}

@article{vrft,
  title={{Visual-RFT}: Visual Reinforcement Fine-Tuning},
  author={Liu, Ziyu and Sun, Zeyi and Zang, Yuhang and Dong, Xiaoyi and Cao, Yuhang and Duan, Haodong and Lin, Dahua and Wang, Jiaqi},
  journal={arXiv preprint arXiv:2503.01785},
  year={2025}
}

@article{lmmr1,
  title={{LMM-R1}: Empowering {3B LMMs} with Strong Reasoning Abilities Through Two-Stage Rule-Based RL},
  author={Peng, Yingzhe and Zhang, Gongrui and Zhang, Miaosen and You, Zhiyuan and Liu, Jie and Zhu, Qipeng and Yang, Kai and Xu, Xingzhong and Geng, Xin and Yang, Xu},
  journal={arXiv preprint arXiv:2503.07536},
  year={2025}
}

@article{r1vl,
  title={{R1-VL}: Learning to Reason with Multimodal Large Language Models via Step-wise Group Relative Policy Optimization},
  author={Zhang, Jingyi and Huang, Jiaxing and Yao, Huanjin and Liu, Shunyu and Zhang, Xikun and Lu, Shijian and Tao, Dacheng},
  journal={arXiv preprint arXiv:2503.12937},
  year={2025}
}

@article{xu2024llavacot,
      title={{LLaVA-CoT}: Let Vision Language Models Reason Step-by-Step},
      author={Guowei Xu and Peng Jin and Hao Li and Yibing Song and Lichao Sun and Li Yuan},
      year={2024},
journal={arXiv preprint arXiv:2411.10440}
}

@inproceedings{li2024synthesize,
  title={Synthesize step-by-step: Tools templates and {LLMs} as data generators for reasoning-based chart {VQA}},
  author={Li, Zhuowan and Jasani, Bhavan and Tang, Peng and Ghadar, Shabnam},
  booktitle={CVPR},
  year={2024}
}

@inproceedings{xia2024geox,
  title={{GeoX}: Geometric Problem Solving Through Unified Formalized Vision-Language Pre-training},
  author={Xia, Renqiu and Li, Mingsheng and Ye, Hancheng and Wu, Wenjie and Zhou, Hongbin and Yuan, Jiakang and Peng, Tianshuo and Cai, Xinyu and Yan, Xiangchao and Wang, Bin and others},
  booktitle={ICLR},
  year={2025}
}

@misc{chen2025r1v,
  author       = {Chen, Liang and Li, Lei and Zhao, Haozhe and Song, Yifan and Vinci},
  title        = {{R1-V}: Reinforcing Super Generalization Ability in Vision-Language Models with Less Than \$3},
  howpublished = {\url{https://github.com/Deep-Agent/R1-V}},
  note         = {Accessed: 2025-02-02},
  year         = {2025}
}

@article{Qwen2.5-VL,
  title={{Qwen2.5-VL} Technical Report},
  author={Bai, Shuai and Chen, Keqin and Liu, Xuejing and Wang, Jialin and Ge, Wenbin and Song, Sibo and Dang, Kai and Wang, Peng and Wang, Shijie and Tang, Jun and Zhong, Humen and Zhu, Yuanzhi and Yang, Mingkun and Li, Zhaohai and Wan, Jianqiang and Wang, Pengfei and Ding, Wei and Fu, Zheren and Xu, Yiheng and Ye, Jiabo and Zhang, Xi and Xie, Tianbao and Cheng, Zesen and Zhang, Hang and Yang, Zhibo and Xu, Haiyang and Lin, Junyang},
  journal={arXiv preprint arXiv:2502.13923},
  year={2025}
}

@article{chen2025sft,
  title={{SFT} or {RL}? An Early Investigation into Training {R1}-Like Reasoning Large Vision-Language Models},
  author={Chen, Hardy and Tu, Haoqin and Wang, Fali and Liu, Hui and Tang, Xianfeng and Du, Xinya and Zhou, Yuyin and Xie, Cihang},
  journal={arXiv preprint arXiv:2504.11468},
  year={2025}
}

@article{chu2025sftmrl,
  title={{SFT} memorizes, {RL} generalizes: A comparative study of foundation model post-training},
  author={Chu, Tianzhe and Zhai, Yuexiang and Yang, Jihan and Tong, Shengbang and Xie, Saining and Schuurmans, Dale and Le, Quoc V and Levine, Sergey and Ma, Yi},
  journal={arXiv preprint arXiv:2501.17161},
  year={2025}
}

@article{guo2025deepseek,
  title={{Deepseek-R1}: Incentivizing reasoning capability in {LLMs} via reinforcement learning},
  author={Guo, Daya and Yang, Dejian and Zhang, Haowei and Song, Junxiao and Zhang, Ruoyu and Xu, Runxin and Zhu, Qihao and Ma, Shirong and Wang, Peiyi and Bi, Xiao and others},
  journal={arXiv preprint arXiv:2501.12948},
  year={2025}
}

@article{marafioti2025smolvlm,
  title={{SmolVLM}: Redefining small and efficient multimodal models},
  author={Marafioti, Andr{\'e}s and Zohar, Orr and Farr{\'e}, Miquel and Noyan, Merve and Bakouch, Elie and Cuenca, Pedro and Zakka, Cyril and Allal, Loubna Ben and Lozhkov, Anton and Tazi, Nouamane and others},
  journal={arXiv preprint arXiv:2504.05299},
  year={2025}
}

@inproceedings{masry-etal-2022-chartqa,
    title = "{C}hart{QA}: A Benchmark for Question Answering about Charts with Visual and Logical Reasoning",
    author = "Masry, Ahmed  and
      Long, Do Xuan  and
      Tan, Jia Qing  and
      Joty, Shafiq  and
      Hoque, Enamul",
    editor = "Muresan, Smaranda  and
      Nakov, Preslav  and
      Villavicencio, Aline",
    booktitle = "Findings of the ACL",
    month = may,
    year = "2022",
}

@inproceedings{mova,
 author = {Zong, Zhuofan and Ma, Bingqi and Shen, Dazhong and Song, Guanglu and Shao, Hao and Jiang, Dongzhi and Li, Hongsheng and Liu, Yu},
 booktitle = {NeurIPS},
 title = {{MoVA}: Adapting Mixture of Vision Experts to Multimodal Context},
 year = {2024}
}

@inproceedings{liu2021slake,
  title={{SLAKE}: A semantically-labeled knowledge-enhanced dataset for medical visual question answering},
  author={Liu, Bo and Zhan, Li-Ming and Xu, Li and Ma, Lin and Yang, Yan and Wu, Xiao-Ming},
  booktitle={ISBI},
  year={2021},
}

@article{BiomedGPT,
  title={A generalist vision--language foundation model for diverse biomedical tasks},
  author={Zhang, Kai and Zhou, Rong and Adhikarla, Eashan and Yan, Zhiling and Liu, Yixin and Yu, Jun and Liu, Zhengliang and Chen, Xun and Davison, Brian D and Ren, Hui and others},
  journal={Nature Medicine},
  pages={1--13},
  year={2024},
  publisher={Nature Publishing Group US New York}
}

@inproceedings{liu-etal-2023-deplot,
    title = "{D}e{P}lot: One-shot visual language reasoning by plot-to-table translation",
    author = "Liu, Fangyu  and
      Eisenschlos, Julian  and
      Piccinno, Francesco  and
      Krichene, Syrine  and
      Pang, Chenxi  and
      Lee, Kenton  and
      Joshi, Mandar  and
      Chen, Wenhu  and
      Collier, Nigel  and
      Altun, Yasemin",
    booktitle = "Findings of the ACL",
    year = "2023",
}

@inproceedings{gao2023gllava,
  title={{G-LLaVA}: Solving geometric problem with multi-modal large language model},
  author={Gao, Jiahui and Pi, Renjie and Zhang, Jipeng and Ye, Jiacheng and Zhong, Wanjun and Wang, Yufei and Hong, Lanqing and Han, Jianhua and Xu, Hang and Li, Zhenguo and others},
  booktitle={ICLR},
  year={2025}
}

@article{zhang2024mathverse,
  title={{MathVerse}: Does Your Multi-modal LLM Truly See the Diagrams in Visual Math {problems?}},
  author={Zhang, Renrui and Jiang, Dongzhi and Zhang, Yichi and Lin, Haokun and Guo, Ziyu and Qiu, Pengshuo and Zhou, Aojun and Lu, Pan and Chang, Kai-Wei and Gao, Peng and others},
  booktitle={ECCV},
  year={2024}
}

@misc{qwen2.5,
    title = {{Qwen2.5}: A Party of Foundation Models},
    url = {https://qwenlm.github.io/blog/qwen2.5/},
    author = {Qwen Team},
    month = {September},
    year = {2024}
}

@article{llavaov,
  	title={{LLaVA-OneVision}: Easy Visual Task Transfer},
  	author={Li, Bo and Zhang, Yuanhan and Guo, Dong and Zhang, Renrui and Li, Feng and Zhang, Hao and Zhang, Kaichen and Li, Yanwei and Liu, Ziwei and Li, Chunyuan},
  	journal={arXiv preprint arXiv:2408.03326},
  	year={2024}
}

@article{intern,
    title={Expanding Performance Boundaries of Open-Source Multimodal Models with Model, Data, and Test-Time Scaling},
    author={Chen, Zhe and Wang, Weiyun and Cao, Yue and Liu, Yangzhou and Gao, Zhangwei and Cui, Erfei and Zhu, Jinguo and Ye, Shenglong and Tian, Hao and Liu, Zhaoyang and others},
    journal={arXiv preprint arXiv:2412.05271},
    year={2024}
  }

@misc{liu2024llavanext,
    title={{LLaVA-NeXT}: Improved reasoning, {OCR}, and world knowledge},
    url={https://llava-vl.github.io/blog/2024-01-30-llava-next/},
    author={Liu, Haotian and Li, Chunyuan and Li, Yuheng and Li, Bo and Zhang, Yuanhan and Shen, Sheng and Lee, Yong Jae},
    month={January},
    year={2024}
}

@article{lai2025med,
  title={{Med-R1}: Reinforcement learning for generalizable medical reasoning in vision-language models},
  author={Lai, Yuxiang and Zhong, Jike and Li, Ming and Zhao, Shitian and Yang, Xiaofeng},
  journal={arXiv preprint arXiv:2503.13939},
  year={2025}
}

@article{zhou2024tinyllava,
  title={{TinyLLaVA}: A framework of small-scale large multimodal models},
  author={Zhou, Baichuan and Hu, Ying and Weng, Xi and Jia, Junlong and Luo, Jie and Liu, Xien and Wu, Ji and Huang, Lei},
  journal={arXiv preprint arXiv:2402.14289},
  year={2024}
}

@misc{openai_o1,
  author       = {{OpenAI}},
  title        = {{Introducing OpenAI o1}},
  howpublished = {\url{https://openai.com/o1/}},
  year         = {2024},
  month        = dec,
  note         = {Accessed: Jun. 21, 2025}
}

@misc{deepseek_r1,
  author       = {{DeepSeek, Inc.}},
  title        = {{DeepSeek-R1 Release}},
  howpublished = {\url{https://api-docs.deepseek.com/news/news250120}},
  year         = {2025},
  month        = jan,
  note         = {Accessed: Jun. 21, 2025}
}

@article{zhang2023multimodal,
  title={Multimodal chain-of-thought reasoning in language models},
  author={Zhang, Zhuosheng and Zhang, Aston and Li, Mu and Zhao, Hai and Karypis, George and Smola, Alex},
  journal={arXiv preprint arXiv:2302.00923},
  year={2023}
}

@article{xia2024chartx,
  title={{ChartX \& ChartVLM}: A versatile benchmark and foundation model for complicated chart reasoning},
  author={Xia, Renqiu and Zhang, Bo and Ye, Hancheng and Yan, Xiangchao and Liu, Qi and Zhou, Hongbin and Chen, Zijun and Ye, Peng and Dou, Min and Shi, Botian and others},
  journal={arXiv preprint arXiv:2402.12185},
  year={2024}
}

@article{yang2025r1one,
  title={{R1-OneVision}: Advancing generalized multimodal reasoning through cross-modal formalization},
  author={Yang, Yi and He, Xiaoxuan and Pan, Hongkun and Jiang, Xiyan and Deng, Yan and Yang, Xingtao and Lu, Haoyu and Yin, Dacheng and Rao, Fengyun and Zhu, Minfeng and others},
  journal={arXiv preprint arXiv:2503.10615},
  year={2025}
}

@inproceedings{
zhai2023investigating,
title={Investigating the Catastrophic Forgetting in Multimodal Large Language Model Fine-Tuning},
author={Yuexiang Zhai and Shengbang Tong and Xiao Li and Mu Cai and Qing Qu and Yong Jae Lee and Yi Ma},
booktitle={CPAL},
year={2023},
}

@inproceedings{liu2023improvedllava,
      title={Improved Baselines with Visual Instruction Tuning}, 
      author={Liu, Haotian and Li, Chunyuan and Li, Yuheng and Lee, Yong Jae},
      booktitle = CVPR,
      year={2024},
}

@article{bai2023qwenvl,
      title={{Qwen-VL}: A Versatile Vision-Language Model for Understanding, Localization, Text Reading, and Beyond}, 
      author={Jinze Bai and Shuai Bai and Shusheng Yang and Shijie Wang and Sinan Tan and Peng Wang and Junyang Lin and Chang Zhou and Jingren Zhou},
      year={2023},
      journal={arXiv preprint arXiv:2308.12966},
}

@misc{korrapati2024moondream,
  author       = {Vik Korrapati},
  title        = {Moondream},
  howpublished = {\url{https://moondream.ai/}},
  year         = {2024},
  note         = {Accessed: 2025-03-27}
}

@article{albalak2022data,
  title={Data-efficiency with a single gpu: An exploration of transfer methods for small language models},
  author={Albalak, Alon and Shrivastava, Akshat and Sankar, Chinnadhurai and Sagar, Adithya and Ross, Mike},
  journal={arXiv preprint arXiv:2210.03871},
  year={2022}
}

@article{ghosh2024exploring,
  title={Exploring the frontier of vision-language models: A survey of current methodologies and future directions},
  author={Ghosh, Akash and Acharya, Arkadeep and Saha, Sriparna and Jain, Vinija and Chadha, Aman},
  journal={arXiv preprint arXiv:2404.07214},
  year={2024}
}

@article{shao2024deepseekmath,
  title={{DeepSeekMath}: Pushing the limits of mathematical reasoning in open language models},
  author={Shao, Zhihong and Wang, Peiyi and Zhu, Qihao and Xu, Runxin and Song, Junxiao and Bi, Xiao and Zhang, Haowei and Zhang, Mingchuan and Li, YK and Wu, Y and others},
  journal={arXiv preprint arXiv:2402.03300},
  year={2024}
}

@misc{GRPO4llava,
  title        = {{GRPO} for {LLaVA}},
  author       = {Yang, Lele and Diao, Muxi and Liang, Kongming and Ma, Zhanyu},
  howpublished = {\url{https://github.com/PRIS-CV/GRPO-for-Llava}},
  year         = {2025}
}

@article{tong2024cambrian,
  title={{Cambrian-1}: A fully open, vision-centric exploration of multimodal {LLMs}},
  author={Tong, Peter and Brown, Ellis and Wu, Penghao and Woo, Sanghyun and IYER, Adithya Jairam Vedagiri and Akula, Sai Charitha and Yang, Shusheng and Yang, Jihan and Middepogu, Manoj and Wang, Ziteng and others},
  journal={Advances in Neural Information Processing Systems},
  volume={37},
  pages={87310--87356},
  year={2024}
}

@article{li2023llavamed,
  title={{LLaVA-Med}: Training a large language-and-vision assistant for biomedicine in one day},
  author={Li, Chunyuan and Wong, Cliff and Zhang, Sheng and Usuyama, Naoto and Liu, Haotian and Yang, Jianwei and Naumann, Tristan and Poon, Hoifung and Gao, Jianfeng},
  journal={Advances in Neural Information Processing Systems},
  volume={36},
  pages={28541--28564},
  year={2023}
}

@inproceedings{duan2024vlmevalkit,
  title={{VLMEvalKit}: An open-source toolkit for evaluating large multi-modality models},
  author={Duan, Haodong and Yang, Junming and Qiao, Yuxuan and Fang, Xinyu and Chen, Lin and Liu, Yuan and Dong, Xiaoyi and Zang, Yuhang and Zhang, Pan and Wang, Jiaqi and others},
  booktitle={ACM MM},
  year={2024}
}

@article{zhang2025policy,
  title={On-policy {RL} meets off-policy experts: Harmonizing supervised fine-tuning and reinforcement learning via dynamic weighting},
  author={Zhang, Wenhao and Xie, Yuexiang and Sun, Yuchang and Chen, Yanxi and Wang, Guoyin and Li, Yaliang and Ding, Bolin and Zhou, Jingren},
  journal={arXiv preprint arXiv:2508.11408},
  year={2025}
}

@article{yan2025learning,
  title={Learning to reason under off-policy guidance},
  author={Yan, Jianhao and Li, Yafu and Hu, Zican and Wang, Zhi and Cui, Ganqu and Qu, Xiaoye and Cheng, Yu and Zhang, Yue},
  journal={arXiv preprint arXiv:2504.14945},
  year={2025}
}

@article{gsm8k,
  title={Training verifiers to solve math word problems},
  author={Cobbe, Karl and Kosaraju, Vineet and Bavarian, Mohammad and Chen, Mark and Jun, Heewoo and Kaiser, Lukasz and Plappert, Matthias and Tworek, Jerry and Hilton, Jacob and Nakano, Reiichiro and others},
  journal={arXiv preprint arXiv:2110.14168},
  year={2021}
}

@article{liu2025better,
  title={Better, Stronger, Faster: Tackling the Trilemma in MLLM-based Segmentation with Simultaneous Textual Mask Prediction},
  author={Liu, Jiazhen and Feng, Mingkuan and Chen, Long},
  journal={arXiv preprint arXiv:2512.00395},
  year={2025}
}

@inproceedings{liu2025phd,
  title={{PhD}: A chatgpt-prompted visual hallucination evaluation dataset},
  author={Liu, Jiazhen and Fu, Yuhan and Xie, Ruobing and Xie, Runquan and Sun, Xingwu and Lian, Fengzong and Kang, Zhanhui and Li, Xirong},
  booktitle={CVPR},
  year={2025}
}

@article{liu2025segmentation,
  title={Segmentation as A Plug-and-Play Capability for Frozen Multimodal LLMs},
  author={Liu, Jiazhen and Chen, Long},
  journal={arXiv preprint arXiv:2510.16785},
  year={2025}
}
\bibliographystyle{iclr2026_conference}
}
\clearpage

\renewcommand{\thesection}{S\arabic{section}}

\renewcommand{\thefigure}{S\arabic{figure}}

\renewcommand{\thetable}{S\arabic{table}}

\let\titleold\title
\renewcommand{\title}[1]{\titleold{#1}\newcommand{\thetitle}{#1}}
\def\maketitlesupplementary{%
   \newpage
   \begin{center}
       {\Large\textbf{\thetitle}}\\[0.5em]
       \large{Supplementary Material}\\[1.0em]
   \end{center}
}
\title{Empowering Small VLMs to Think with Dynamic Memorization and Exploration}
\setcounter{page}{1}
\maketitlesupplementary

\setcounter{section}{0}
\setcounter{figure}{0}
\setcounter{table}{0}
\setcounter{equation}{0}

In the supplementary materials, we report:
\begin{itemize}

\item LLM instructions used for constructing vision supervision (\S \ref{sec:inst});
\item Detailed experimental setup and additional experimental results (\S \ref{sec:exp});
\item Showcases of SVLMs trained via \model performing on medical VQA, chart understanding, and geometry problem solving (\S\ref{sec:showcase});
\end{itemize}

\section{LLM Instructions for Vision Supervision}
\label{sec:inst}
The instructions for constructing $I_c$, the visual refiner, and the visual checker are listed as follows. 

\subsection{Instructions for Extracting Visual Elements}

$I_c$ is primarily derived from two sources: ground truth captions, and the outputs from specialized tools such as the chart-parsing model Deplot. Prompt~\ref{lst:fact-aokvqa} is employed to extract visual elements from captions.








\begin{lstlisting}[caption=\textbf{Automated Visual Fact Extraction}, label={lst:fact-aokvqa}]
You are a helpful assistant that analyzes images and provides visual facts.
Your response MUST be a single, valid JSON object.
The JSON object should contain:
1. "description": A detailed and accurate description of the image.
2. "objects": A list of key objects, including their name, attributes, and approximate position in the image.

Example format:
{
  "description": "A person riding a bicycle on a city street.... (detailed description here)",
  "objects": [
    {"name": "person", "attributes": ["wearing helmet", "blue shirt"], "position": "center"},
    {"name": "bicycle", "attributes": ["red", "mountain bike"], "position": "center"},
    {"name": "street", "attributes": ["asphalt", "wet"], "position": "bottom"}
  ]

Analyze the attached image and provide the visual facts in the required JSON format.
For context, the user will be asked this question about the image (do not answer the question, just use it for context):
"{question}"
}
\end{lstlisting}

\subsection{Instructions for Visual Refiner}

The visual refiner primarily takes thinking templates from a template pool and $I_c$ as input to construct the ground truth for SFT.

\begin{lstlisting}[caption=\textbf{Ground-truth construction for Chart Understanding SFT}, label={lst:vr-chart-precise}]
Given:
<IC>: the data of an image
<Q>: a question
<A>: a reference answer
<T>: a writing template

Goal:
Transform the visual information in <IC> into a textualized data description and incorporate it into a smooth, natural explanation that reasons why the correct answer to <Q> is <A>, using the format and tone defined by <T>.

Example:
<IC>: [
  {"object": "bar", "attributes": ["~120k", "Q4"], "label": "Product A"},
  {"object": "bar", "attributes": ["~150k", "Q4"], "label": "Product B"},
  {"object": "bar", "attributes": ["~90k", "Q4"], "label": "Product C"},
  {"title": "Quarterly Revenue"}
]
<Q>: Which product has the highest revenue in Q4?
<A>: product b
<T>:
    [Extraction] Extract the relevant data points from <IC> that are needed to answer <Q>.
    [Calculation] Analyze and compare the extracted data to determine the correct result.
    [Conclusion] Based on the analysis, state why <A> is the correct answer to <Q>.

<Output>:
    [Extraction] The image shows a chart comparing three items-Product A, Product B, and Product C-in Q4. The bar heights indicate their revenues: Product A at about 120k, Product B at about 150k, and Product C at about 90k.
    [Calculation] Comparing these values, Product B's revenue (150k) is greater than both Product A's (120k) and Product C's (90k).
    [Conclusion] Since Product B has the highest revenue among the three, the correct answer to the question "Which product has the highest revenue in Q4?" is product b.

Now, according to the requirements and the examples above, convert my input into the target reasoning text. Please give me the result directly without any explanation or description.

<IC>: %s
<Q>: %s
<A>: %s
<T>: %s
<Output>:
\end{lstlisting}

Prompts for the other domains follow a similar design.


\subsection{Instructions for Visual Checker}

The visual checker is primarily responsible for scoring the thinking trace of responses generated in the GRPO process. It evaluates these traces with reference to exemplars, based on their fluency and the degree to which the mentioned visual elements align with $I_c$. Prompts for the other domains follow a similar design.

\begin{lstlisting}[caption=\textbf{Scoring generations during GRPO for Chart Understanding }, label={lst:vc-chart-precise}]
Given
<IC>: the data of an image
<Q>: a question
<A>: a reference answer
<R>: a reasoning text

Goal:
Assess whether <R> correctly and reasonably uses visible data in <IC> to justify that the correct answer to <Q> is <A>. Rate the quality as low / medium / high according to:
(a) low: Does not use data from <IC> at all, or the language is not fluent/natural, or it fails to indicate the answer to <Q> is <A>.
(b) medium: Uses data from <IC> and is written fluently, but the reasoning is overly brief or insufficiently clear.
(c) high: Uses data from <IC> and is written fluently; the reasoning progresses step by step with depth, each step is correct and reasonable; the data from <IC> appears exactly where it should; overall, the reasoning text provides very strong support that the answer to <Q> is <A>.

Example:
<IC>: [
  {"object": "bar", "attributes": ["~120k", "Q4"], "label": "Product A"},
  {"object": "bar", "attributes": ["~150k", "Q4"], "label": "Product B"},
  {"object": "bar", "attributes": ["~90k", "Q4"], "label": "Product C"},
  {"title": "Quarterly Revenue"}
]
<Q>: Which product has the highest revenue in Q4?
<A>: product b
<R>:
    [Extraction] Reads Q4 bar heights: A ~120k, B ~150k, C ~90k.
    [Calculation] Compares values: B > A and B > C.
    [Conclusion] Therefore, Product B is highest, matching the answer "product b".

<Output>: medium

According to the requirements and examples above, score the input into three categories. Please give me the result directly without any explanation or description.

<IC>: %s
<Q>: %s
<A>: %s
<R>: %s
<Output>: 
\end{lstlisting}

\section{Experimental Details and Extra Results}
\label{sec:exp}

\subsection{Training and Test Setting}
\begin{figure}[htbp]
\centering
\setlength{\tabcolsep}{6pt}
\renewcommand{\arraystretch}{1.12}
\begin{minipage}[t]{0.31\linewidth}
\small
\textbf{Medical (SLAKE).}\\
\includegraphics[width=\linewidth]{Template/figures/medical.jpg}
\textbf{\textsf{Q}}: What is the MR weighting in this image?\\
\textbf{w/o refiner}:\\
Based on signal variations typical of this image, the most plausible answer is \textbf{T2}.\\
\textbf{w/ refiner}:\\
\textit{Analysis:} On T2, \textcolor{darkgreen}{fluid} is bright; \textcolor{darkgreen}{bone/air} and fast flow are dark.\\
\textit{Extraction:} In this \textcolor{darkgreen}{axial slice}, \textcolor{darkgreen}{intraluminal fluid} is bright; \textcolor{darkgreen}{vessels} low signal; \textcolor{darkgreen}{ribs/wall} dark.\\
\textit{Conclusion:} Pattern matches \textbf{T2}.\\
\end{minipage}\hfill
\begin{minipage}[t]{0.31\linewidth}
\small
\textbf{Chart (ChartQA).}\\
\includegraphics[width=\linewidth]{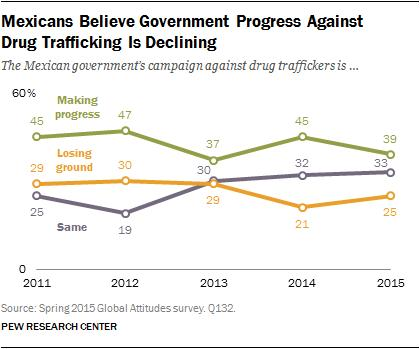}\\
\textbf{\textsf{Q}}: What is the sum of \emph{making progress}, \emph{losing ground}, and \emph{same} in 2011?\\
\textbf{w/o refiner}:\\
To answer the question, we first locate the relevant part of the chart for 2011, then extract the values for the three series and sum them up to obtain \textbf{99}.\\
\textbf{w/ refiner}:\\
\textit{Extraction:} Read 2011 values: \textcolor{darkgreen}{Making progress = 45}, \textcolor{darkgreen}{Losing ground = 29}, \textcolor{darkgreen}{Same = 25}.\\
\textit{Calculation:} 45 + 29 + 25 = 99.\\
\textit{Conclusion:} The sum is \textbf{99}.
\end{minipage}\hfill
\begin{minipage}[t]{0.31\linewidth}
\small
\textbf{Math (Geo170k).}\\
\includegraphics[width=\linewidth]{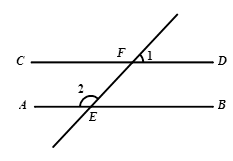}\\
\textbf{\textsf{Q}}: If angle 1 is \(35^\circ\), what is angle 2? Choices: A: \(55^\circ\), B: \(135^\circ\), C: \(145^\circ\), D: \(155^\circ\).\\
\textbf{w/o refiner}:\\
 Since angle 1 is 35 degree, angle 2 is 180 - 35 = 145 degree using a supplementary relationship. Choose \textbf{C}.\\
\textbf{w/ refiner}:\\
\textit{Extraction:} Two \textcolor{darkgreen}{parallel} lines CD and AB with a \textcolor{darkgreen}{transversal}; angle 1 is 35 degree, and angle 2 is on the \textcolor{darkgreen}{same side}.\\
\textit{Calculation:} \textcolor{darkgreen}{Same-side interior angles are supplementary}, so angle 2 = 180 - 35 = 145 degree.\\
\textit{Conclusion:} Answer: \textbf{C}.
\end{minipage}
\caption{\textbf{Comparison of ground-truth responses before and after refinement.} 
Compared to the original ground-truth, the refiner injects richer visual elements and enforces a more structured organization, thereby reducing the learning burden for SVLMs.}
\label{fig:visual}
\end{figure}

First, we provide the statistical information for the training and testing phases of our experiments in the Tab. \ref{tab:set}. The training dataset for each domain consists of only a few thousand samples. In addition, Fig.~\ref{fig:visual} visualizes a comparison between the ground-truth responses produced by the refiner and the original ground-truth, showing that the refined versions are noticeably more structured and place greater emphasis on intermediate values.

\begin{table}[htbp]
\raa{1.2}
\centering
\caption{\textbf{Training and testing setup.} \model empowers thinking capabilities based on small training sets.}
\resizebox{0.8\linewidth}{!}{
\begin{tabular}{@{}llllr@{}}
\toprule
\textbf{Domain} & \textbf{Training set} & \textbf{\#Training samples} & \textbf{Source of $I_c$}  & \textbf{Testset}  \\
\midrule
Medical VQA & SLAKE-Train & 4,919 & 	
BiomedGPT & SLAKE-Test \\
Chart Understanding & ChartQA-Train & 4,576 & DePlot & ChartQA-Test \\
Geometry Solving & Geo170K & 6,417 & Collected& MathVerse \\
\bottomrule
\end{tabular}
}

\label{tab:set}
\end{table}

\subsection{Extra Results}

We also report additional experimental content, including the discussion on training strategies and data organization formats, as well as a comparative analysis with other similar methods that integrate SFT and RL.

\begin{wraptable}{r}{0.4\linewidth}
\centering
\captionof{table}{\textbf{Two-stage training on ChartQA}. \texttt{Rel-corr} denotes the relaxed-correctness metric. \(I_c\) indicates whether an explicit image-content field is supervised (\ding{51} yes; \ding{55} no).}
\label{tab:smolvlm-chartqa-sft}
    \resizebox{0.65\linewidth}{!}{
\begin{tabular}{@{}lrr@{}}
\toprule
\textbf{Model} & \textbf{\(I_c\)} & \textbf{Rel-corr} \\
\midrule
SmolVLM      & \ding{51} & 64.32 \\
SmolVLM      & \ding{55} & 60.09 \\
LLaVA-OV-S   & \ding{51} & 63.62 \\
LLaVA-OV-S   & \ding{55} & 52.90 \\
\bottomrule
\end{tabular}
}
\end{wraptable}

Specifically, (1) we first demonstrate the importance of constructing vision supervision, which proves essential for training SVLMs to produce grounded thinking traces. (2) We then examine the impact of structured versus open-ended output formats on thinking performance. (3) Furthermore, to validate our earlier observation that SVLMs are prone to converging to local optima, we present performance across different training epochs, showing that SFT training saturates after only one epoch. (4) We provide a detailed comparison with alternative methods that integrate SFT and RL. (5) Finally, we extend our evaluation to stronger base models and pure textual domains, and (6) validate the quality of generated thinking traces through human evaluation.

\begin{figure}[ht]
  \centering
  \includegraphics[width=\linewidth]{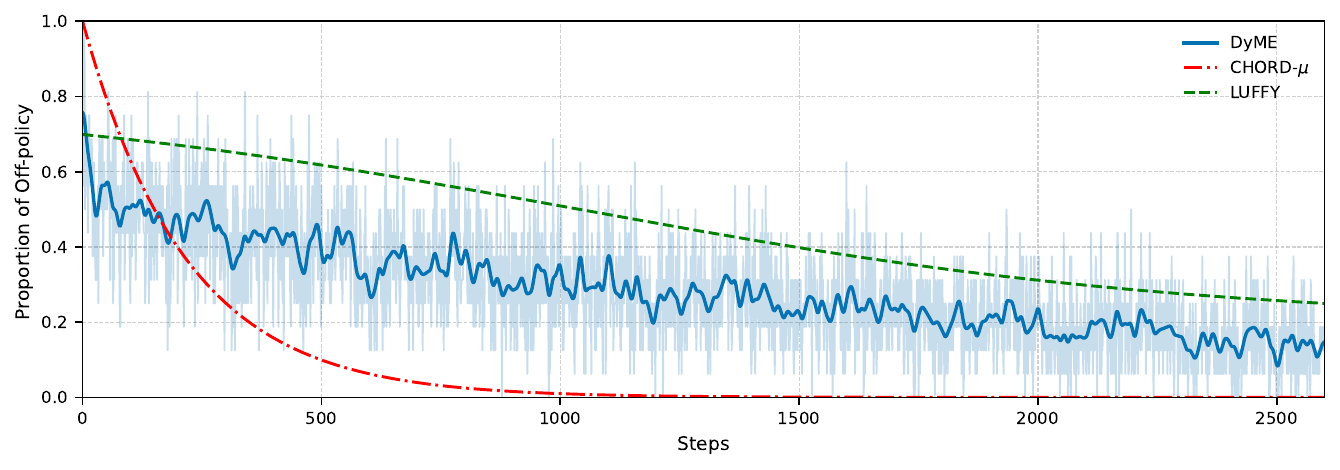}
\caption{\textbf{Relative off-policy influence during training.}
Each curve is normalized to its initial value for comparability. \model measures $\mathrm{SFT}/(\mathrm{SFT}+\mathrm{RL})$ (raw in lighter tone, Gaussian-smoothed in darker tone), \textsc{CHORD-$\mu$} tracks the global weight $\mu(t)$, and \textsc{LUFFY} adopts a policy-shaping proxy $\mathbb{E}[f(\pi_\theta(a))]$ with $f(x)=\tfrac{x}{x+\gamma}$. All methods reveal the shift from off-policy guidance to on-policy optimization, albeit with distinct decay dynamics.}
  \label{fig:offpolicy-transition-comparison}
\end{figure}

\begin{table}[h]
\centering
\small
\caption{\textbf{Effect of templated output across models and tasks.} 
\checkmark denotes fixed-template output, whereas \ding{55} denotes free-form generation.}
\begin{tabular}{lccc}
\toprule
\textbf{Model} & \textbf{Template} & \textbf{Chart} & \textbf{Medical} \\
\midrule
SmolVLM       & \checkmark & 60.10 & 59.38 \\
SmolVLM       & \ding{55}  & 59.24 & 56.13 \\
LLaVA-OV-S    & \checkmark & 52.87 & 74.52 \\
LLaVA-OV-S      & \ding{55}  & 50.86 & 72.64 \\
\bottomrule
\end{tabular}
\label{tab:template_effect}
\end{table}

\textbf{(1) Intermediate values matter.}
As shown in Table~\ref{tab:smolvlm-chartqa-sft}, we report the effect of applying two-stage training with visual supervision on SmolVLM and LLaVA-OV-S. Incorporating visual supervision significantly improves the best performance achieved during training, despite certain instabilities, thereby validating its critical role for SVLMs. This effect is further illustrated in Fig.~\ref{fig:visual}, where visual supervision compels SVLMs to generate intermediate reasoning enriched with visual elements, which make a clear contribution to the final answer.

\textbf{(2) Structured thinking alleviates the learning burden of SVLMs.}
Table~\ref{tab:template_effect} reports the performance gap between training with structured thinking ground-truth and with unconstrained ground-truth. While open-ended exploration is often beneficial for LVLMs, the limited capacity of SVLMs makes unconstrained exploration less effective, as it tends to be aimless and increases the learning burden. Given that SVLMs are designed for task-specific rather than general-purpose scenarios, employing tailored thinking templates for each task proves more suitable and yields better performance. For instance, SmolVLM achieves $60.10$ \emph{vs.}~$59.24$ on ChartQA and $59.38$ \emph{vs.}$56.13$ on Medical VQA, with LLaVA-OV-S exhibiting similar gains.

\begin{wraptable}{r}{0.4\linewidth}
  \centering
  \caption{\textbf{SVLM performance saturates after a single training epoch.}
  Score is domain-specific: chart domain uses \texttt{Rel-corr}, while the medical domain uses the average of accuracy and recall values.}
  \label{tab:smolvlm-chartqa-sft-means}
\resizebox{0.9\linewidth}{!}{
  \begin{tabular}{@{}ll
                  r
                  r
                  @{}}
    \toprule
    \textbf{Model} & \textbf{Domain} & {\textbf{Epoch}} & {\textbf{Score}} \\
    \midrule
    \multirow{3}{*}{LLaVA-OV-S} & \multirow{3}{*}{Chart}   & 1  & 60.70 \\
                                &                           & 5  & 60.44 \\
                                &                           & 10 & 60.12 \\
    \midrule
    \multirow{6}{*}{SmolVLM}    & \multirow{3}{*}{Chart}    & 1  & 60.22 \\
                                &                           & 5  & 63.21 \\
                                &                           & 10 & 62.22 \\
    \cmidrule(l){2-4}
                                & \multirow{3}{*}{Medical}  & 1  & 71.73 \\
                                &                           & 5  & 71.80 \\
                                &                           & 10 & 72.05 \\
    \bottomrule
  \end{tabular}
  }
\end{wraptable}

\textbf{(3) Comparison between annealed SFT loss and \model.}
As shown in Fig.~\ref{fig:offpolicy-transition-comparison}, we compare the {relative SFT (off-policy) influence} across training steps for three approaches: \model, CHORD~\citep{zhang2025policy}, and LUFFY~\citep{yan2025learning}. For \model and CHORD, the curves represent the normalized weight of the SFT loss at each step, while for LUFFY the curve reflects the trajectory of SFT gradient shaping as a function of prediction probability (which generally correlates with training steps). These curves highlight the dynamic nature of \model. Because of the extremely limited capacity of SVLMs, their learning patterns can shift significantly even between adjacent steps, leading to rapid forgetting of previously acquired modes. Unlike CHORD, which relies on a smooth annealing schedule that decays quickly and is ill-suited to such small models, \model assigns weights directly based on model outputs.

This produces a highly dynamic and irregular decay, better accommodating the instability of SVLMs.
LUFFY adopts a shaping function $f(x) = \tfrac{x}{x+\gamma}$ ($\gamma{=}0.1$), which also induces a dynamic decay with probability but remains heuristic and may not align well with the local-optimum tendency of SVLMs. 
Overall, \model is explicitly tailored for SVLMs, whereas CHORD and LUFFY may be more appropriate for stronger base models, reflecting complementary strengths.

\textbf{(4) SVLMs converge rapidly.}
Table~\ref{tab:smolvlm-chartqa-sft-means} shows that SVLMs converge extremely quickly: performance after only one epoch is comparable to, or even exceeds, that after ten epochs (\eg, LLaVA-OV-S achieves 60.70 \emph{vs.}~60.12 on the Chart domain). 
This indicates that the very limited capacity of SVLMs makes them prone to overfitting to local optima. It also substantiates our earlier claim that such rapid convergence leaves only a narrow window for balancing SFT and RL, making it difficult to achieve the trade-off through empirical hyperparameter tuning. Consequently, static fusion methods are unsuitable for SVLMs.

{To ensure a rigorous comparison, we further report the full learning trajectories of baselines in Table~\ref{tab:learning-trajectories}. We evaluated the Two-stage baseline (with and without KL penalty) and SFT across multiple epochs (1, 3, 5, 10) to capture their peak performance. The results confirm that even with optimal stopping, the baselines consistently underperform \model, which achieves superior results in a single training run without the need for epoch selection.

\begin{table}[t]
\centering
\scriptsize
\caption{{\textbf{Detailed learning trajectories demonstrating rigorous tuning.} We report the performance across multiple settings to show their full learning trajectories. Two-stage baselines include variants with and without KL penalties to ensure optimal performance is captured.}}
\label{tab:learning-trajectories}
\setlength{\tabcolsep}{4pt} 
\resizebox{\linewidth}{!}{%
\begin{tabular}{llcc}
\toprule
\textbf{Data Quality} & \textbf{Method} & \textbf{Performance across epochs (1, 3, 5, 10)} & \textbf{Best perf.} \\
\midrule
\multirow{4}{*}{Low}
  & \model (ours, pure)      & \emph{Report final score directly}                         & \textbf{61.9} \\
  & SFT                      & $43.1 \rightarrow 47.9 \rightarrow 50.0 \rightarrow 50.5$  & 50.5           \\
  & Two-stage                & $57.6 \rightarrow 52.7 \rightarrow 50.8 \rightarrow 50.7$  & 57.6           \\
  & Two-stage (w/ KL)        & $54.2 \rightarrow 55.4 \rightarrow 52.6 \rightarrow 54.2$  & 55.4           \\
\midrule
\multirow{4}{*}{Medium}
  & \model (ours, pure)      & \emph{Report final score directly}                         & \textbf{64.9} \\
  & SFT                      & $53.6 \rightarrow 56.5 \rightarrow 57.8 \rightarrow 56.4$  & 57.8           \\
  & Two-stage                & $59.9 \rightarrow 52.8 \rightarrow 53.0 \rightarrow 53.1$  & 59.9           \\
  & Two-stage (w/ KL)        & $59.0 \rightarrow 60.6 \rightarrow 60.6 \rightarrow 60.8$  & 60.8           \\
\midrule
\multirow{4}{*}{High}
  & \model (ours, pure)      & \emph{Report final score directly}                         & \textbf{68.5} \\
  & SFT                      & $58.2 \rightarrow 59.1 \rightarrow 61.0 \rightarrow 61.6$  & 61.6           \\
  & Two-stage                & $51.6 \rightarrow 54.0 \rightarrow 54.5 \rightarrow 54.4$  & 54.5           \\
  & Two-stage (w/ KL)        & $61.7 \rightarrow 60.9 \rightarrow 62.7 \rightarrow 61.8$  & 62.7           \\
\bottomrule
\end{tabular}
}
\end{table}

\textbf{(5) Generality across complex reasoning and pure text.} 
To demonstrate the scalability of \model, we applied it to two new domains without modifying the core algorithm: {Physical Reasoning} (A-OKVQA) and \textbf{Pure Text Reasoning} (GSM8K).

\begin{itemize}[leftmargin=*]
    \item \textbf{Physical Reasoning (A-OKVQA):} We addressed the challenge of open-ended visual reasoning by testing on A-OKVQA. We used Qwen2.5-VL-7B to automatically generate Visual Facts using the prompt defined in \S\ref{lst:fact-aokvqa} (e.g., \textit{``man, wearing a light blue and white shirt...''}). As shown in Table~\ref{tab:dyme-generalization}, \model achieved a massive gain of {+18.8\%} ($54.2\% \to 73.0\%$), proving that the method scales effortlessly to tasks requiring world knowledge and commonsense.
    \item \textbf{Pure Text Reasoning (GSM8K):} In pure text domains, the ``Visual Fact'' extraction step is naturally skipped. On the GSM8K math benchmark, \model improved Qwen2.5-0.5B from $49.5\%$ to {$55.3\%$}, demonstrating that the paradigm generalizes even when ``vision'' is absent.
\end{itemize}

These results, combined with the ChartQA improvements on the stronger Qwen2.5-VL-7B model, confirm that \model is not limited by the extraction step. By leveraging off-the-shelf LVLMs to automate visual fact generation, the framework is immediately applicable to diverse visual and textual domains.
}

{\textbf{Limitations on Abstract Visuals.} We acknowledge that the VS module may face challenges in scenarios where ``Visual Facts'' are intrinsically difficult to define or extract, such as memes (relying on irony or cultural context) or highly abstract non-commonsense reasoning. However, our primary objective is to empower SVLMs for practical, real-world production tasks (\eg, chart processing, medical diagnostics, geometric solving). In these structured and semi-structured domains where SVLMs are most commonly deployed, Visual Facts are well-defined and \model proves highly effective.}

\begin{table}[t]
\centering
\scriptsize
\caption{{\textbf{Generality of \model across New Domains.} We demonstrate performance gains on Complex Scenes (A-OKVQA), Pure Text (GSM8K), and with stronger base models (Qwen2.5-VL-7B). Baselines for text use standard GRPO.}}
\label{tab:dyme-generalization}
\setlength{\tabcolsep}{4pt} 
\resizebox{\linewidth}{!}{%
\begin{tabular}{llllcc}
\toprule
\textbf{Domain} & \textbf{Task} & \textbf{Base Model} & \textbf{Method} & \textbf{Baseline (\%)} & \textbf{\model (\%)} \\
\midrule
\textbf{World Knowledge} & A-OKVQA & LLaVA-OV-S & Two-stage & 54.2 & \textbf{73.0 (+18.8)} \\
\textbf{Pure Text}       & GSM8K  & Qwen2.5-0.5B & GRPO      & 49.5 & \textbf{55.3 (+5.8)} \\
\textbf{New LVLM}        & ChartQA & Qwen2.5-VL-7B & SFT      & 87.3 & \textbf{89.6 (+2.3)} \\
\bottomrule
\end{tabular}
}
\end{table}

{\textbf{(6) Human evaluation of CoT quality.} 
Automatic metrics like relaxed accuracy do not fully reflect the quality of the reasoning process. To verify whether \model generates genuinely better thinking traces, we conducted a human evaluation on 100 randomly sampled instances from ChartQA. Annotators judged the validity of the generated CoT based on its logical coherence and grounding. As shown in Table~\ref{tab:cot-human-eval-chartqa}, \model produces traces that are slightly more concise (shorter length) but significantly more valid (validity rate $\sim$70\%) compared to the Two-stage baseline ($\sim$30-40\%). This confirms that \model effectively mitigates the generation of ``pseudo thinking traces'' that plague standard SFT/Two-stage training.}

\begin{table}[htbp]
\centering
\scriptsize
\caption{\textbf{Human evaluation of CoT quality on ChartQA.}}
\label{tab:cot-human-eval-chartqa}
\setlength{\tabcolsep}{4pt} 
\resizebox{0.8\linewidth}{!}{%
\begin{tabular}{llrr}
\toprule
\textbf{Base Model} & \textbf{Method} & \textbf{Avg. CoT Length} & \textbf{Human Eval (Valid \%)} \\
\midrule
\multirow{2}{*}{\textbf{LLaVA-OV-S}}
  & Two-stage & $\sim$76.3 Words & 31\% \\
  & DyME      & $\sim$69.7 Words & \textbf{68\%} \\
\midrule
\multirow{2}{*}{\textbf{SmolVLM}}
  & Two-stage & $\sim$84.5 Words & 40\% \\
  & DyME      & $\sim$75.4 Words & \textbf{72\%} \\
\bottomrule
\end{tabular}
}
\end{table}

\section{Showcases} 
\label{sec:showcase}

{Before presenting the model outputs, we first illustrate the data quality definitions used in our Algorithmic Validation (Section 4.1 of the main paper). Table~\ref{tab:cot-quality-showcase} showcases examples of Low (Undesigned), Medium (Standard), and High (Premium) quality Chain-of-Thought supervision for the same question. This visualizes the significant gap in structure and detail that \model must bridge when trained on non-premium data.} Furthermore, Table~\ref{tab:aokvqa-quality-showcase} illustrates the comprehensive format of our supervision data, encompassing the input image, the associated question, the extracted visual facts, and the ground-truth response.

\begin{figure}[htb]
  \centering

  \begin{subfigure}{\linewidth}
    \centering
    \includegraphics[width=0.8\linewidth]{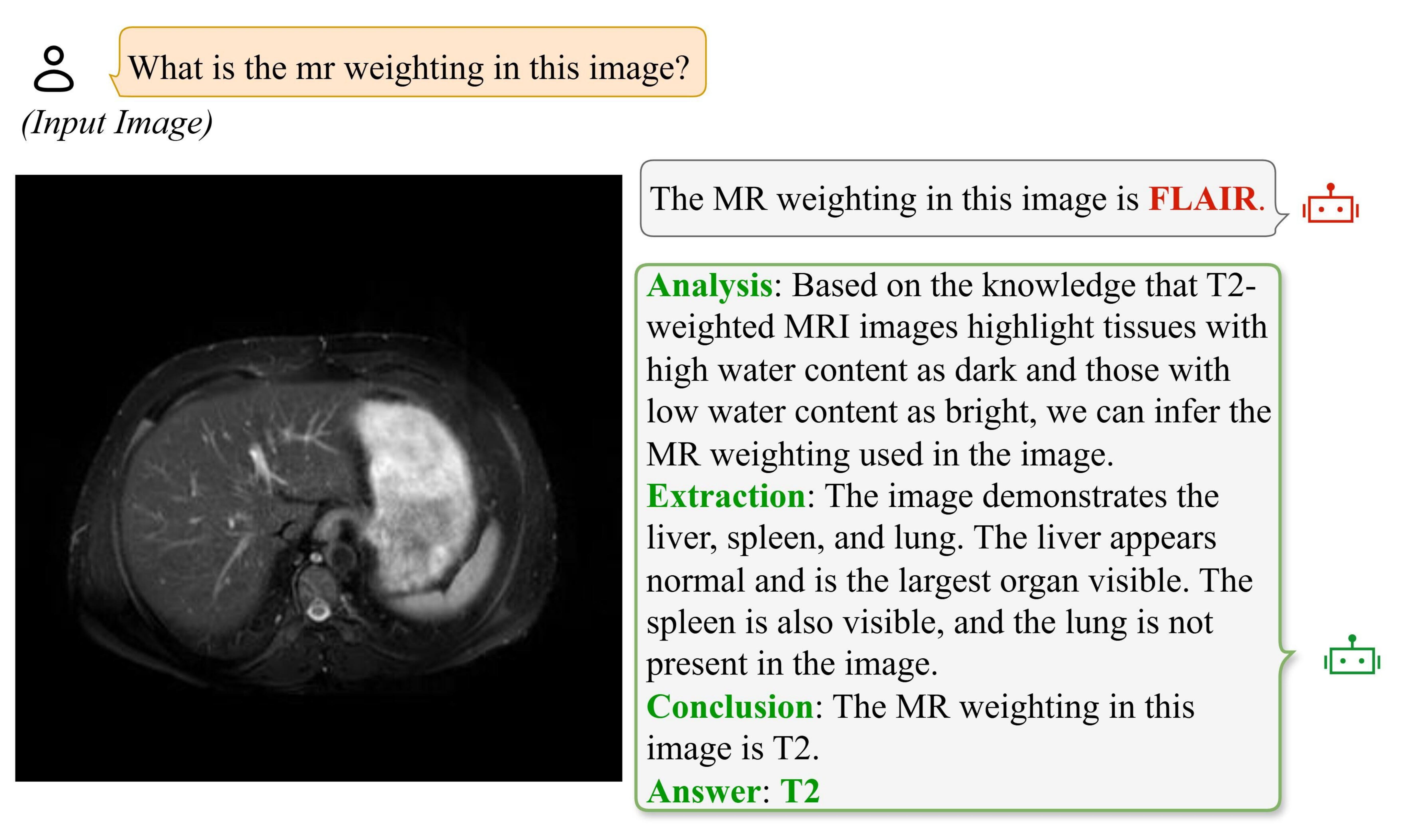}
    \caption{Showcase of Medical VQA}
  \end{subfigure}

      \begin{subfigure}{\linewidth}
    \centering
    \includegraphics[width=0.8\linewidth]{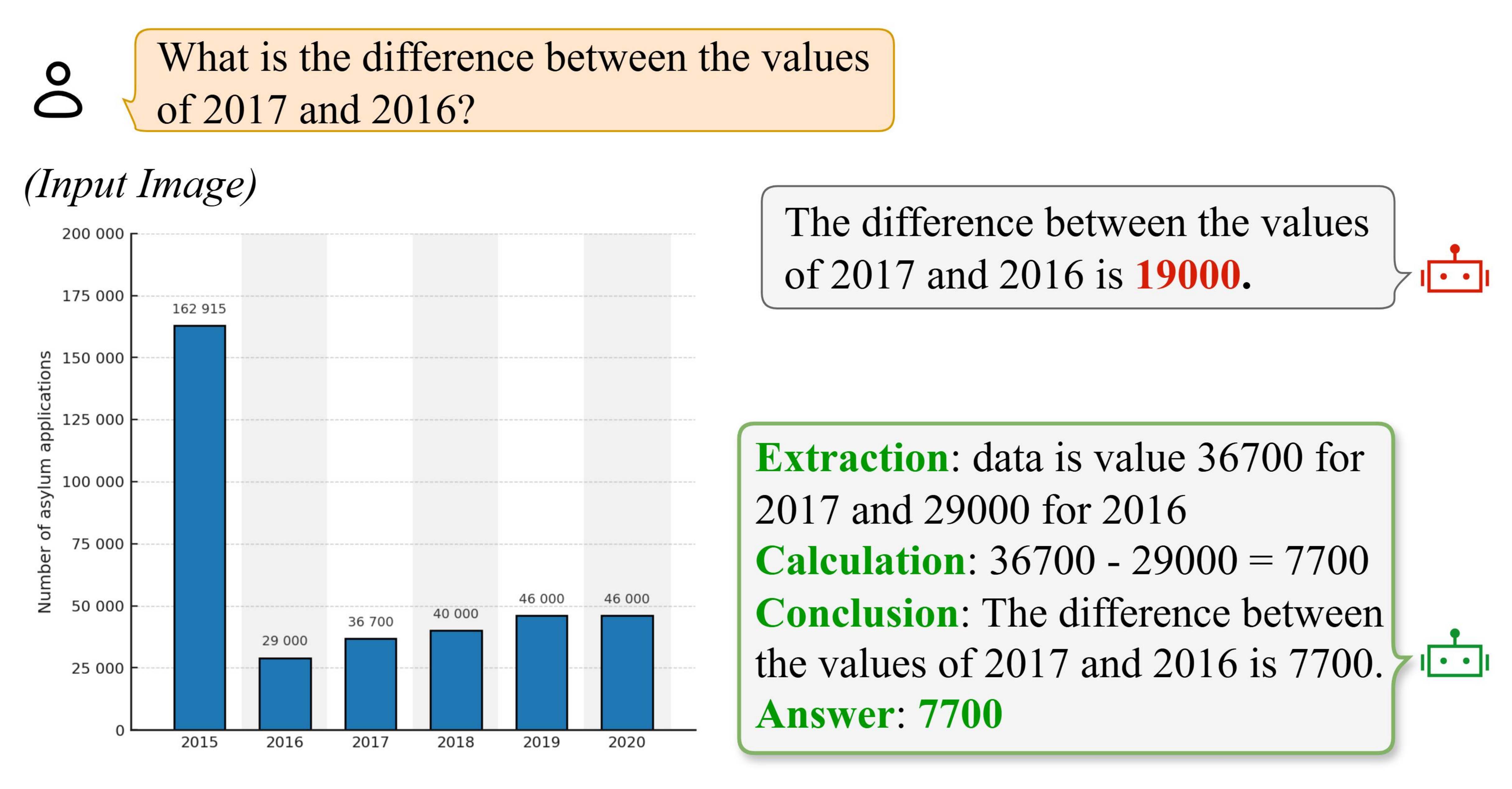}
    \caption{Showcase of Chart Understanding}
  \end{subfigure}

  \begin{subfigure}{\linewidth}
    \centering
    \includegraphics[width=0.8\linewidth]{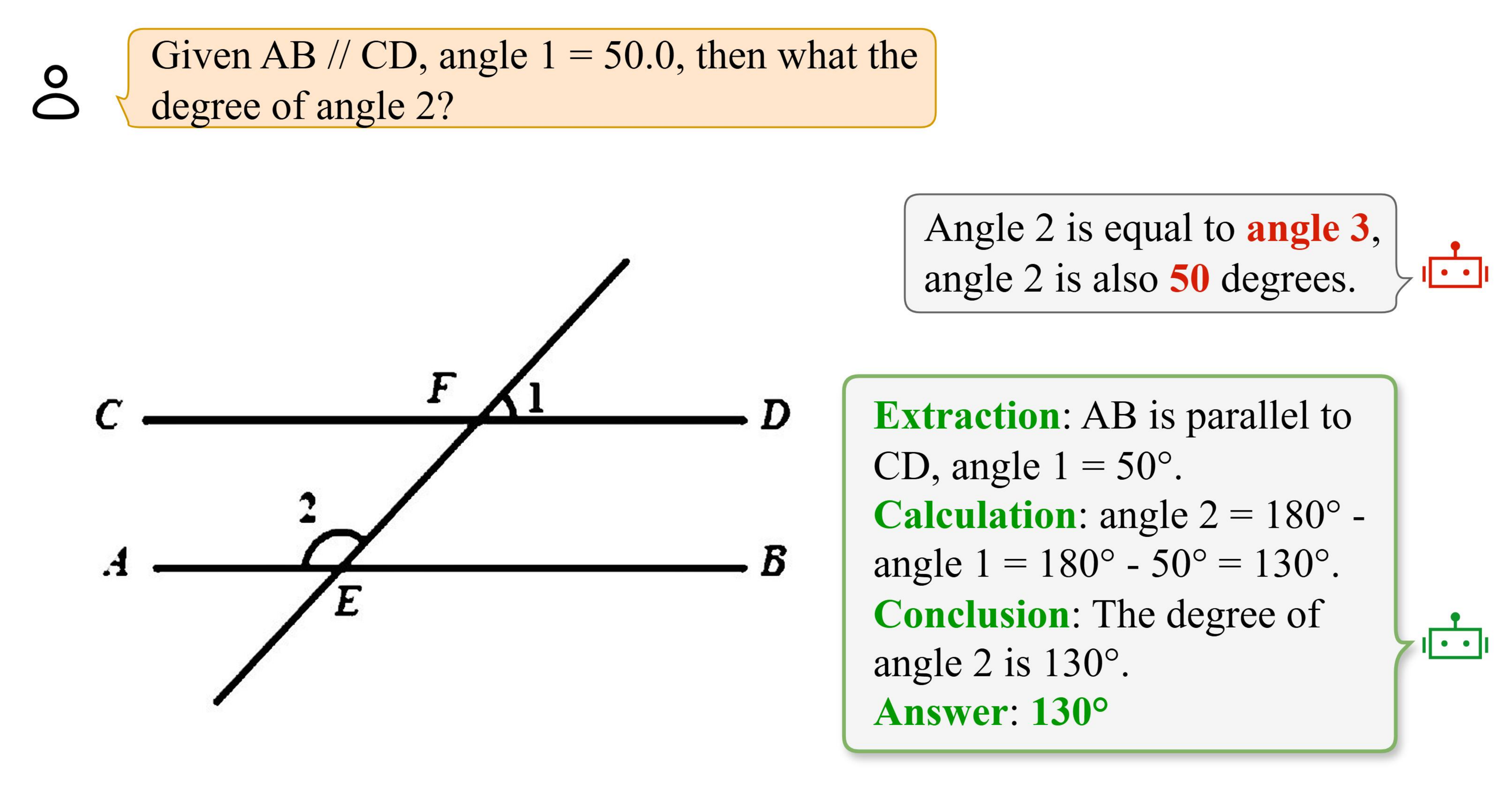}
    \caption{Showcase of Geometry Solving}
  \end{subfigure}

  \caption{\textbf{Showcases of SmolVLM.} The SVLM originally produces hallucinated answers (red), while the \model-trained model generates structured thinking traces (green) that incorporate grounded values, effectively improving the performance.}
  \label{fig:showcases1}
\end{figure}

We present dialogue instances of SmolVLM, LLaVA-OV-S, and InternVL2-S, which were trained with \model\ in our experiments, on tasks in the domains of medical VQA, chart understanding, and geometry. As shown in Fig~\ref{fig:showcases1} to Fig~\ref{fig:showcases3}, all models trained with \model\ demonstrate the ability to generate effective thinking traces with accurate intermediate values (in green), which play a crucial role in reaching the final correct answer.

The thinking processes of these models are relatively fixed and template-based. This is a result of our vision supervision module taking effect, as SVLMs should not be overly broad and general; otherwise, their exploration can easily diverge instead of converging.

\begin{table}[t]
\small
\centering
\caption{\textbf{Showcase of chain-of-thought (CoT) supervision with different quality for ChartQA.}}
\label{tab:cot-quality-showcase}
\resizebox{0.9\linewidth}{!}{
\begin{tabular}{p{0.15\textwidth} >{\raggedright\arraybackslash}p{0.85\textwidth}}
\toprule
\textbf{Type} & \textbf{Content} \\
\midrule
Question & When does the unfavorable view reach the peak? \\
& \begin{minipage}[t]{\linewidth}
    \includegraphics[width=0.4\linewidth]{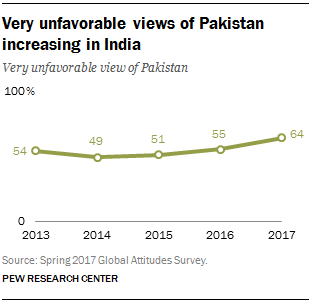}
  \end{minipage} \\
\midrule
High-quality CoT
& \begin{minipage}[t]{\linewidth}\footnotesize
\texttt{<SUMMARY>} I inspect the line graph showing very unfavorable views of Pakistan in India over time, and identify the year where the percentage peaks by locating the maximum point on the curve. \texttt{</SUMMARY>}\\[0.35em]
\texttt{<CAPTION>} Line graph titled ``Very unfavorable views of Pakistan increasing in India''. The y-axis shows the share of respondents with a very unfavorable view (0--100\%). The x-axis shows years 2013--2017. Marked values: 54\% (2013), 49\% (2014), 51\% (2015), 55\% (2016), and 64\% (2017). The series exhibits an overall upward trend and peaks in 2017. \texttt{</CAPTION>}\\[0.35em]
\texttt{<REASONING>} The peak corresponds to the highest percentage on the y-axis. Among the annotated years, 2017 has the largest value (64\%), higher than 2013--2016. Therefore, the peak occurs in 2017. \texttt{</REASONING>}\\[0.35em]
\texttt{<CONCLUSION>} 2017 \texttt{</CONCLUSION>}
\end{minipage}\\[0.6em]
\midrule
Medium-quality CoT
& \begin{minipage}[t]{\linewidth}\footnotesize
\textbf{Goal:} Determine the year when the unfavorable view reaches its maximum.\\
\textbf{Observation:} The values for each year are: 2013: 0, 2014: 0, 2015: 0, 2016: 55, and 2017: 64.\\
\textbf{Reasoning:} Comparing these values, the largest value is 64, which appears in 2017.\\
\textbf{Conclusion:} The unfavorable view reaches its peak in 2017.
\end{minipage}\\[0.6em]
\midrule
Low-quality CoT
& \begin{minipage}[t]{\linewidth}\footnotesize
I'm trying to figure out the year when the unfavorable view reaches its highest point. The values stay low until 2016, where they rise to 55, and then climb further to 64 in 2017. This makes 2017 the year when the unfavorable view reaches its maximum.
\end{minipage}\\
\bottomrule
\end{tabular}
}
\end{table}

\begin{table}[t]
\small
\centering
\caption{\textbf{Showcase of a complete CoT training example.}}
\label{tab:aokvqa-quality-showcase}
\resizebox{0.9\linewidth}{!}{
\begin{tabular}{p{0.15\textwidth} >{\raggedright\arraybackslash}p{0.85\textwidth}}
\toprule
\textbf{Type} & \textbf{Content} \\
\midrule
Question & What is the man by the bags awaiting? \\
& \begin{minipage}[t]{\linewidth}
    \includegraphics[width=0.4\linewidth]{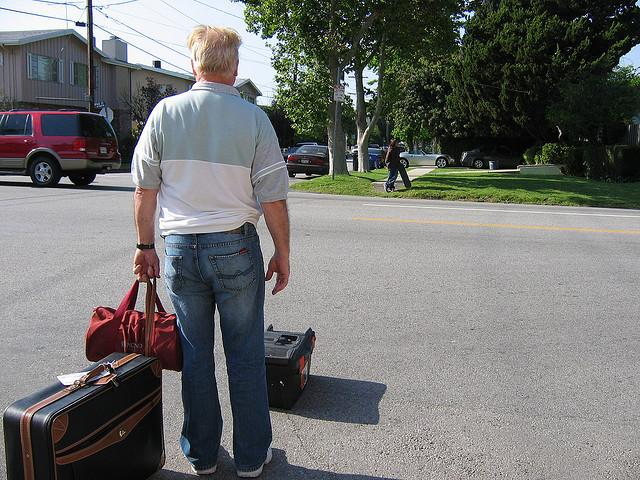}
  \end{minipage} \\
\midrule
Visual Fact
& \begin{minipage}[t]{\linewidth}\footnotesize
\{
  "description": "A man stands in the middle of a residential street, facing away from the camera. He holds a red bag and pulls a wheeled black suitcase, with another black suitcase placed nearby. Houses, parked cars, and trees line the background, suggesting he is waiting to cross or be picked up.",
  "objects": [
    {
      "name": "man",
      "attributes": ["light blue and white shirt", "blue jeans", "carrying a red bag", "pulling a wheeled suitcase"],
      "position": "center"
    },
    {
      "name": "red bag",
      "attributes": ["held by the man"],
      "position": "left of the man"
    },
    {
      "name": "black suitcase",
      "attributes": ["wheeled", "being pulled"],
      "position": "near the man's feet"
    },
    {
      "name": "black suitcase",
      "attributes": ["lying on the ground"],
      "position": "near the man"
    },
    {
      "name": "street",
      "attributes": ["asphalt", "urban residential"],
      "position": "foreground"
    },
    {
      "name": "houses",
      "attributes": ["residential homes"],
      "position": "background left"
    },
    {
      "name": "parked cars",
      "attributes": ["red SUV", "other vehicles"],
      "position": "background center and left"
    },
    {
      "name": "trees",
      "attributes": ["green foliage"],
      "position": "background right"
    }
  ]
\}
\end{minipage}\\
\midrule
GT Response
& \begin{minipage}[t]{\linewidth}\footnotesize
A train would not appear on a residential street, and the man’s luggage suggests he is waiting for transportation rather than a delivery. The nearby skateboarder is not interacting with him, leaving a cab as the most plausible option.
\end{minipage}\\[0.6em]
\bottomrule
\end{tabular}
}
\end{table}

\begin{figure}[p]
  \centering

  \begin{subfigure}{\linewidth}
    \centering
    \includegraphics[width=0.8\linewidth]{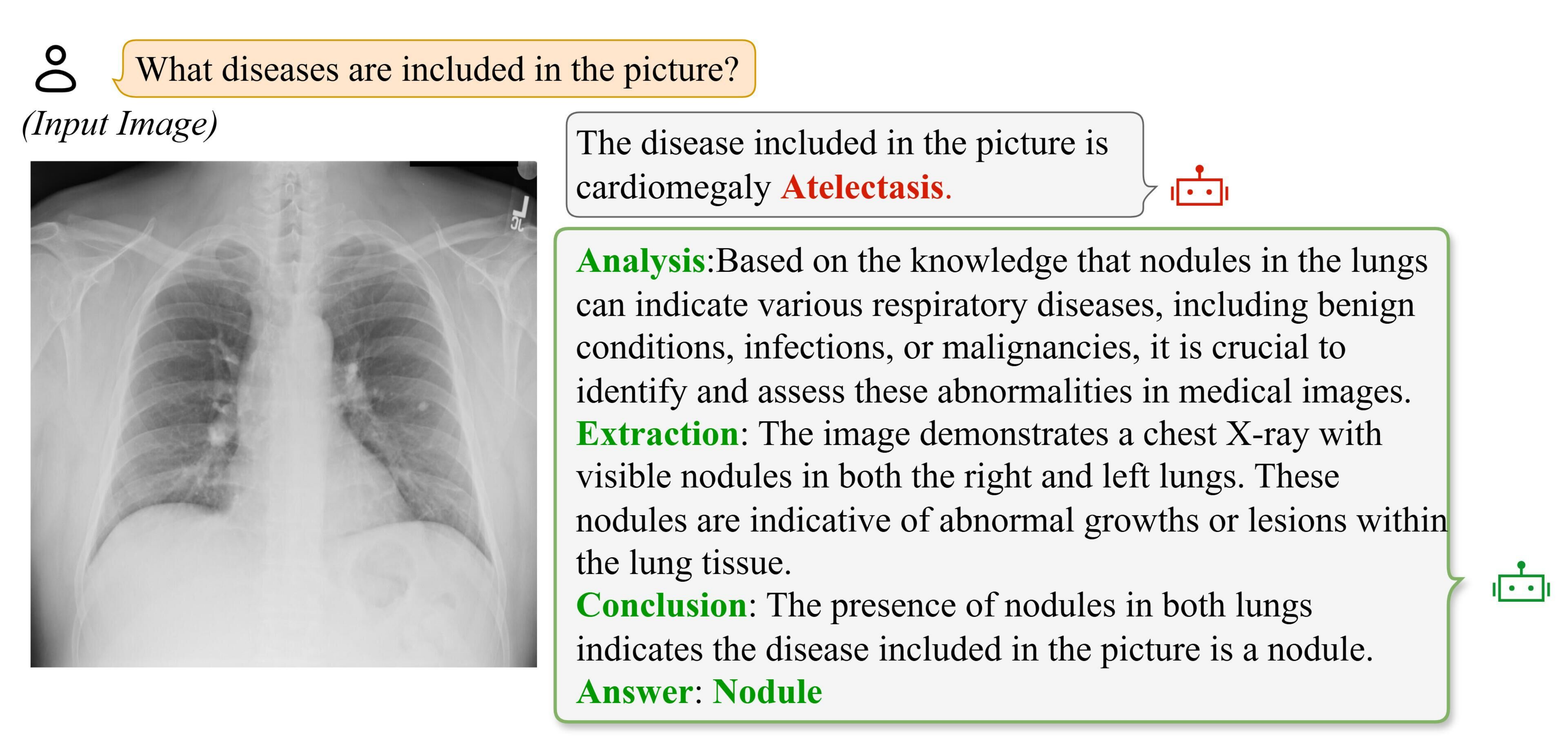}
    \caption{Showcase of Medical VQA}
  \end{subfigure}

      \begin{subfigure}{\linewidth}
    \centering
    \includegraphics[width=0.8\linewidth]{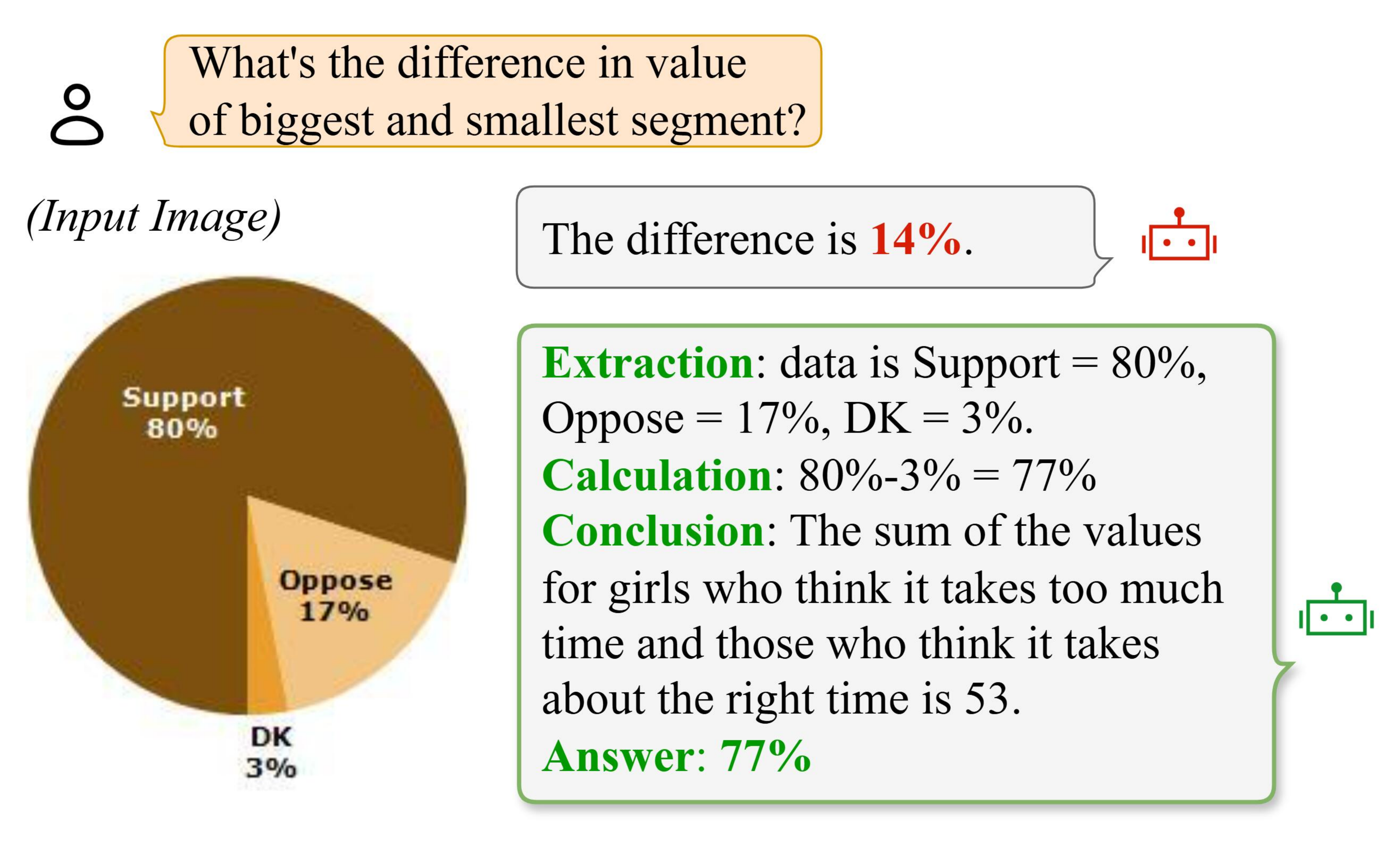}
    \caption{Showcase of Chart Understanding}
  \end{subfigure}

  \begin{subfigure}{\linewidth}
    \centering
    \includegraphics[width=0.8\linewidth]{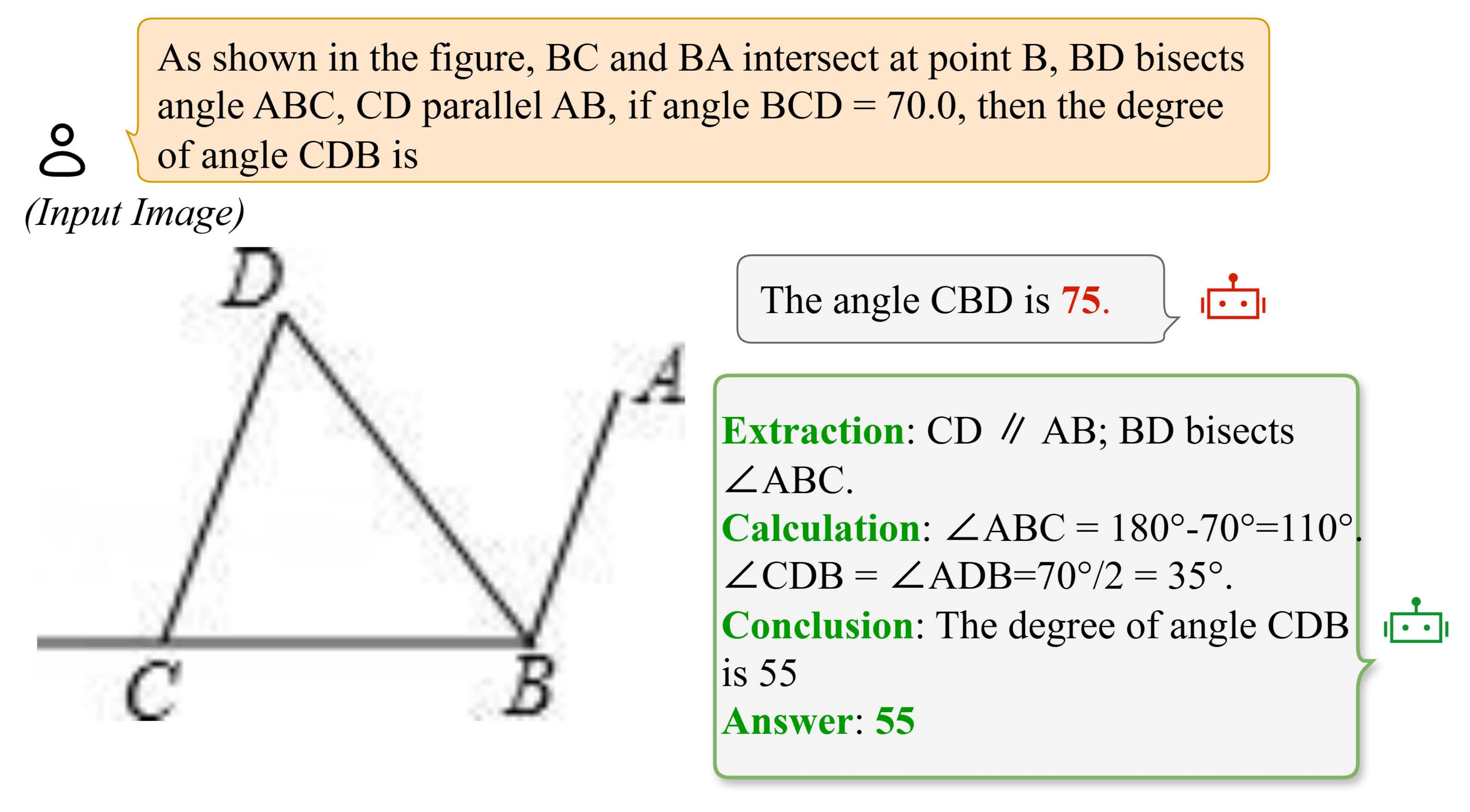}
    \caption{Showcase of Geometry Solving}
  \end{subfigure}

  \caption{\textbf{Showcases of InternVL2-S.} The SVLM originally produces hallucinated answers (red), while the \model-trained model generates structured thinking traces (green) that incorporate grounded values, effectively improving the performance.}
  \label{fig:showcases2}
\end{figure}

\begin{figure}[p]
  \centering
  \begin{subfigure}{\linewidth}
    \centering
    \includegraphics[width=0.8\linewidth]{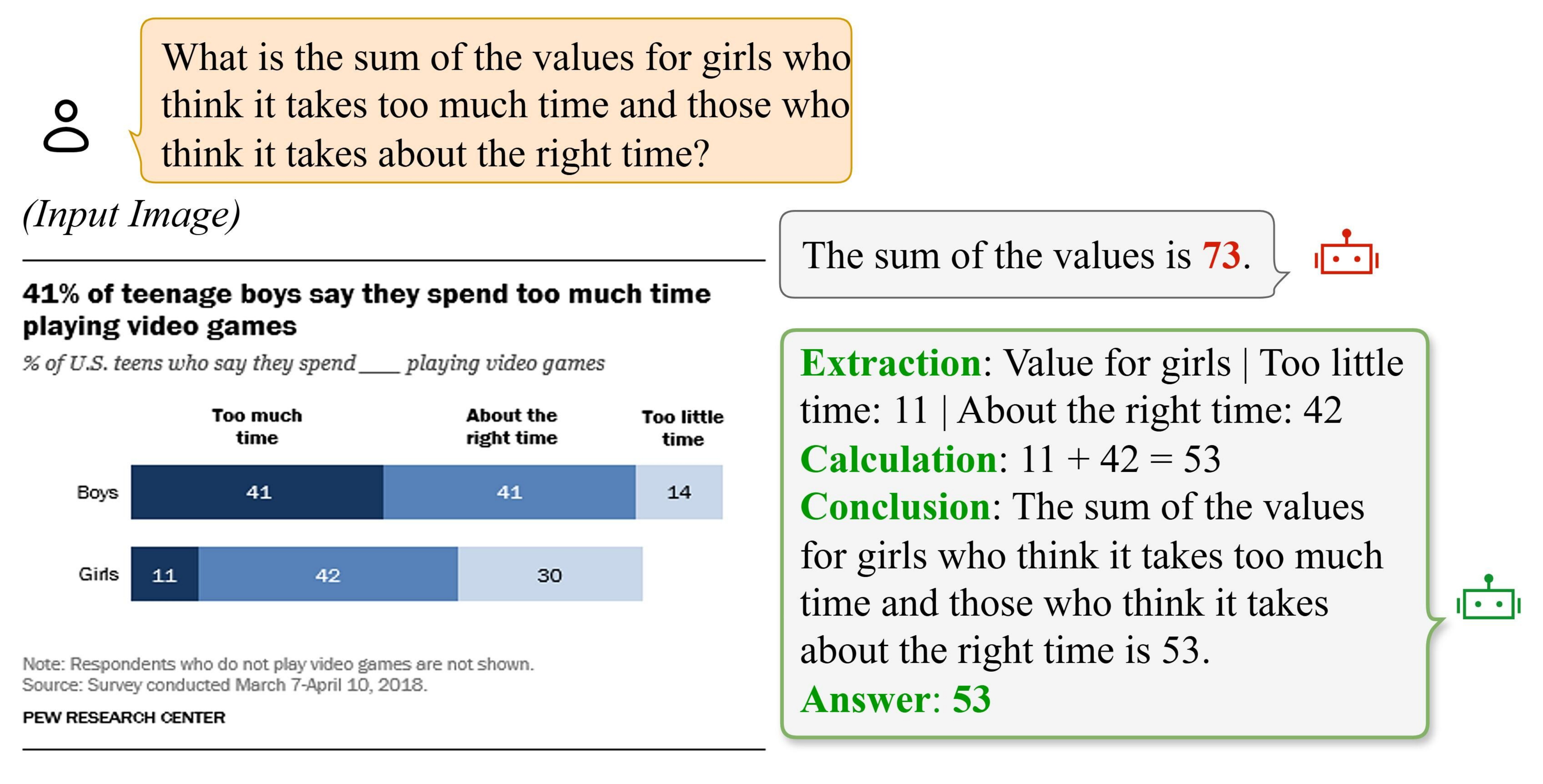}
    \caption{Showcase of Chart Understanding}
  \end{subfigure}

      \begin{subfigure}{\linewidth}
    \centering
    \includegraphics[width=0.8\linewidth]{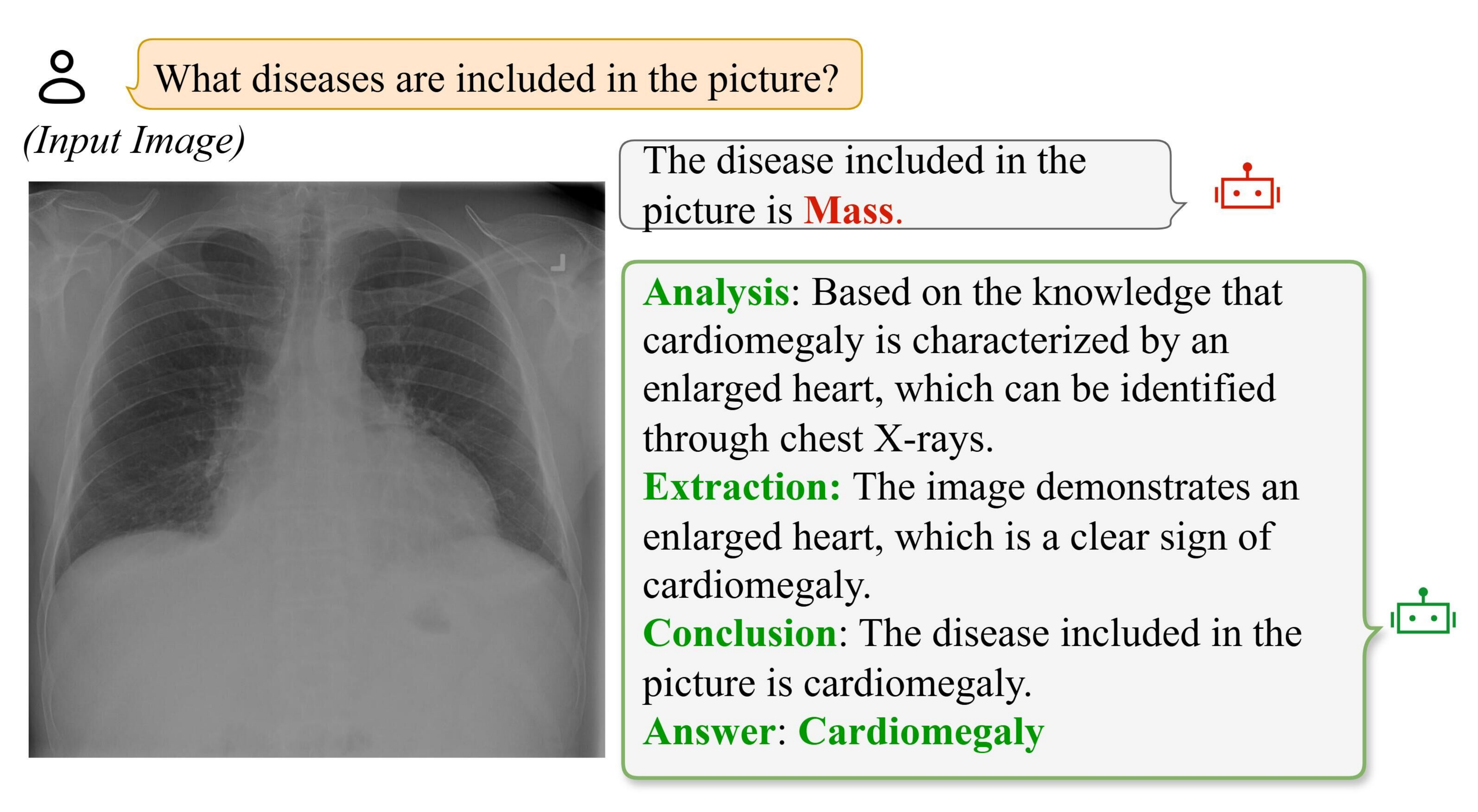}
    \caption{Showcase of Medical VQA}
  \end{subfigure}

  \begin{subfigure}{\linewidth}
    \centering
    \includegraphics[width=0.8\linewidth]{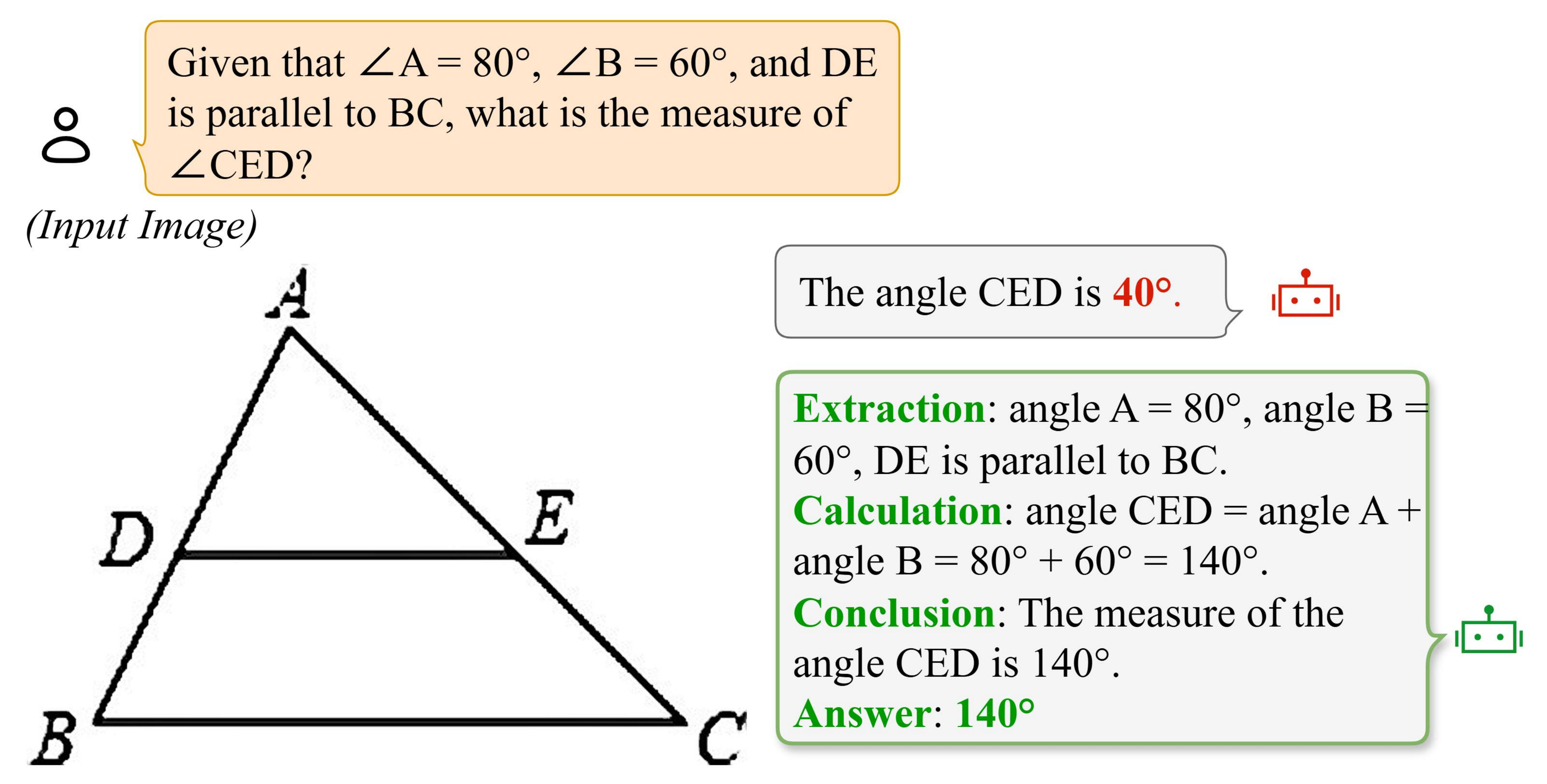}
    \caption{Showcase of Geometry Solving}
  \end{subfigure}

  \caption{\textbf{Showcases of LLaVA-OV-S.} The SVLM originally produces hallucinated answers (red), while the \model-trained model generates structured thinking traces (green) that incorporate grounded values, effectively improving the performance.}
  \label{fig:showcases3}
\end{figure}

\end{document}